%% file: 0-main.tex
\pdfoutput=1

\documentclass[11pt]{article}

\usepackage{acl}
\usepackage{amsmath} 
\usepackage{times}
\usepackage{latexsym}

\usepackage[T1]{fontenc}

\usepackage[utf8]{inputenc}

\usepackage{microtype}

\usepackage{inconsolata}

\usepackage{graphicx}

\usepackage{multirow}
\usepackage{verbatim}
\usepackage{longtable}
\usepackage{lscape}
\usepackage{hyperref}
\usepackage{graphicx}
\usepackage{url}
\usepackage{enumitem}
\usepackage{subcaption}
\usepackage{booktabs}
\usepackage{longtable}
\usepackage{array}
\usepackage{tabularx}  
\usepackage{lipsum} 

\title{What Matters in Evaluating Book-Length Stories? \\A Systematic Study of Long Story Evaluation}

\makeatletter
\def\thanks#1{\protected@xdef\@thanks{\@thanks
        \protect\footnotetext{#1}}}
\makeatother
\author{Dingyi Yang, \bf Qin Jin*\thanks{ *Corresponding Author.}\\
         Renmin University of China \\ 
         \texttt{\{yangdingyi,qjin\}@ruc.edu.cn} \\
         }

\begin{document}
\maketitle                
\begin{abstract}
In this work, we conduct systematic research in a challenging area: the automatic evaluation of book-length stories (>100K tokens). Our study focuses on two key questions: (1) understanding which evaluation aspects matter most to readers, and (2) exploring effective methods for evaluating lengthy stories. We introduce the first large-scale benchmark, \textbf{LongStoryEval}, comprising 600 newly published books with an average length of 121K tokens (maximum 397K). Each book includes its average rating and multiple reader reviews, presented as critiques organized by evaluation aspects. By analyzing all user-mentioned aspects, we propose an \textit{evaluation criteria structure} and conduct experiments to identify the most significant aspects among the 8 top-level criteria. For evaluation methods, we compare the effectiveness of three types: \textit{aggregation-based, incremental-updated}, and  \textit{summary-based} evaluations. Our findings reveal that aggregation- and summary-based evaluations perform better, with the former excelling in detail assessment and the latter offering greater efficiency. Building on these insights, we further propose \textbf{NovelCritique}, an 8B model that leverages the efficient summary-based framework to review and score stories across specified aspects. NovelCritique outperforms commercial models like GPT-4o in aligning with human evaluations. Our datasets and codes are available at \url{https://github.com/DingyiYang/LongStoryEval}.


\end{abstract}

\input{sections/1-intro}
\input{sections/2-related_works}

\input{sections/3-benchmark}
\input{sections/4-method}

\input{sections/5-experiments}
\input{sections/6-results}

\input{sections/7-conclusion}

\section*{Acknowledgements}
We thank all reviewers for their insightful comments and suggestions. 
This work was partially supported by the Beijing Natural Science Foundation (No. L233008).

\bibliography{custom}

\appendix

\input{sections/appendix}

\end{document}

%% file: sections/1-intro.tex
\section{Introduction}


\begin{figure}[t]
\centering
    \includegraphics[width=.48\textwidth]{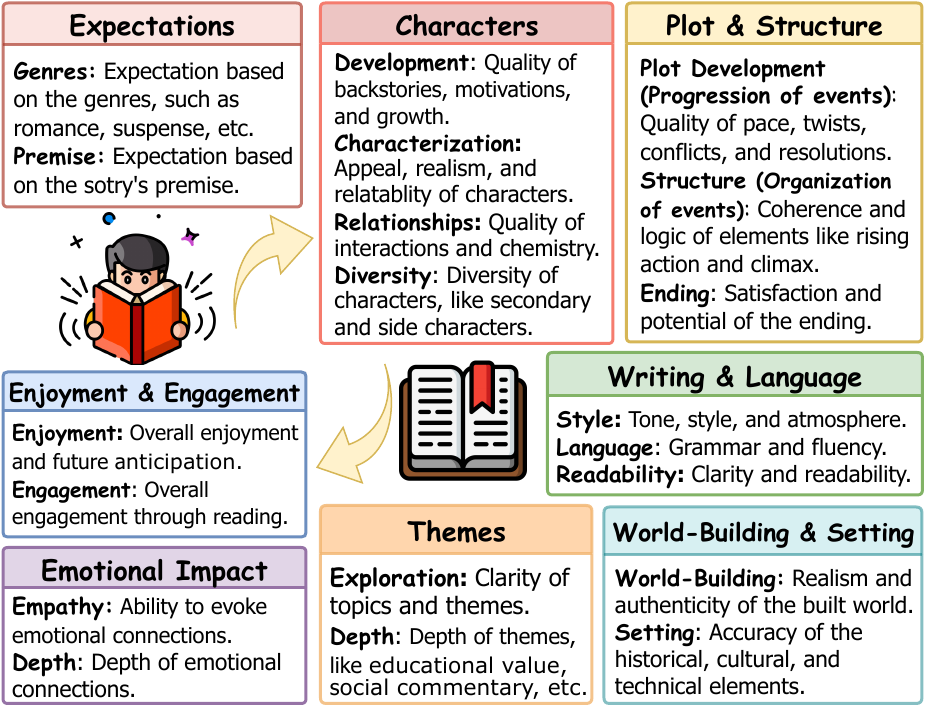}

    \caption{Our proposed \textit{evaluation criteria structure} and the reading process: A reader approaches a book with initial \textbf{expectations} based on its genres and premise. The story unfolds through \textbf{character}-driven \textbf{plots}, revealing its \textbf{themes} and \textbf{world-building} through the author's \textbf{writing}. Through reading, the reader experiences \textbf{enjoyment}, \textbf{engagement}, and \textbf{emotional impact}, and determines whether this book meets the expectations.}
    \vspace{-5px}
    \label{fig:evaluation_aspects}
\end{figure}

\vspace{-7px}

\textbf{Automatic Story Evaluation} involves providing critiques and ratings to assess the quality of human-written or machine-generated stories. This process is crucial for recommendation systems or offering constructive feedback for improvement.  Unlike simpler evaluation tasks that focus on fluency and accuracy (e.g., machine translation), story evaluation demands a comprehensive assessment, based on diverse human-centered criteria ~\cite{chhun2022hanna}. While recent advances have improved the evaluation of short stories ~\citep{guan2020union,guan2021openmeva,chhun2022hanna}, particularly with the aid of large language models (LLMs)~\cite{jiang2023tigerscore,xie2023deltascore}, the evaluation of book-length stories (exceeding 100K tokens) remains significantly underexplored.

Evaluating book-length stories poses three major challenges: (1) \textbf{Data Annotation Constraints}: Human evaluation, while the gold standard, is time-intensive and cognitively demanding. As shown in Table \ref{table:eval_dataset}, existing story evaluation benchmarks focus on shorter texts (100-2,000 tokens). Scaling human annotations for stories exceeding 100K tokens is impractical. 
(2) \textbf{Inconsistent Evaluation Criteria}: Most prior works rely on predefined criteria for evaluation, but there is no universal standard. 
Evaluation criteria vary across studies and often fail to reflect actual reader preferences. Our work aims to explore what \textbf{real readers} value in lengthy stories. 
(3) \textbf{Long Story Processing}: Book-length stories often exceed the 128K-token context limit of most LLMs, posing challenges for effective evaluation. Even within this limit, processing such long contexts remains challenging for models. 
Identifying efficient evaluation strategies for lengthy stories is therefore critical. 

To address these challenges, we collect ratings and reviews for 600 newly published lengthy novels from online readers. To completely avoid data contamination issues \cite{chang2023booookscore} that might affect our experimental analysis, none of these books were included in the training data of evaluated LLMs. The raw review data is sourced from GoodReads\footnote{\url{https://www.goodreads.com}}, the largest book review platform. 
Using LLMs, we extract over 1000 reader-mentioned evaluation aspects, analyze the most frequent ones, and organize them into a \textit{hierarchical criteria structure} (Figure \ref{fig:evaluation_aspects}).
We further compare three types of processing methods for lengthy story evaluation: \textit{aggregation-based}, \textit{incremental-updated}, and \textit{summary-based}. Additionally, we introduce \textit{NovelCritique}, a specialized model for reviewing and scoring lengthy stories across specified aspects,  which 
demonstrates superior alignment with human evaluations compared to commercial models. 

\begin{figure*}[t]
    \includegraphics[width=1\textwidth]{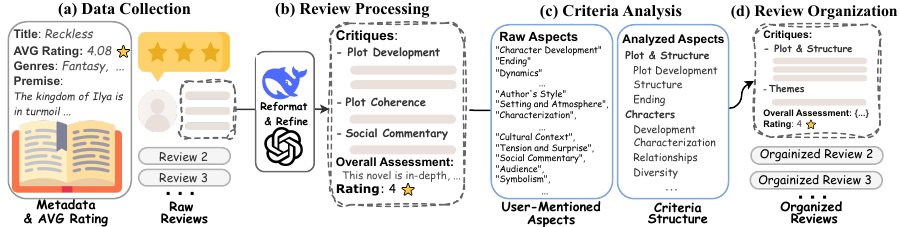}
    \caption{Our data construction process (\S \ref{sec:benchmark}). }
    \vspace{-5px}
    \label{fig:data_construction}
    \centering
\end{figure*}

Our contributions are summarized as follows:
\begin{itemize}[leftmargin=*]
\vspace{-3px}
\item \textbf{LongStoryEval: A benchmark for lengthy story evaluation.} 
We introduce a large-scale benchmark comprising 600 books (published between 2024 and January 2025), with average rating scores and 340K reader reviews.
Raw reviews are converted into structured critiques, overall assessments, and ratings, as shown in Figure \ref{fig:data_construction} (d). Metadata, including book details (e.g., title, genres, premise) and reviewer profiles, is provided to facilitate future research.
\vspace{-3px}
\item \textbf{A hierarchical structure of evaluation criteria, and analysis of significant aspects.} By analyzing all aspects raised by real readers, we develop a hierarchical evaluation criteria structure with 8 main aspects and 20 sub-aspects (Figure \ref{fig:evaluation_aspects}; Table \ref{table:detailed_criteria}). Our experiments reveal that \textit{plot} and \textit{characters} are the most influential objective aspects, while subjective aspects --- \textit{emotional impact}, \textit{overall enjoyment \& engagement}, and \textit{expectation fulfillment} --- are also critical to overall ratings.
\vspace{-3px}
\item \textbf{Explorations on effective methods for lengthy story evaluation.}
Among the three types of lengthy story processing methods, \textit{aggregation-} and \textit{summary-based} evaluations perform best. Our findings suggest that the most cost-efficient method involves generating a concise summary and averaging multiple summary-based evaluation results. Further experimental analysis is provided in \S \ref{sec:analysis}.

\item \textbf{\textit{NovelCritique}: A specialized model for lengthy story evaluation.} We propose NovelCritique, an 8B model capable of reviewing and scoring lengthy stories across specified aspects. It outperforms commercial LLMs such as GPT-4o in aligning with human ratings.

\end{itemize}

%% file: sections/2-related_works.tex
\vspace{-3px}
\section{Related Works}\label{sec:related_work}
\paragraph{Story Evaluation.} Story generation is a creative and open-ended task, making it more appropriate to explore metrics based on specific human standards. Traditional lexical-based metrics such as BLEU \cite{papineni2002bleu} and METEOR \cite{banerjee2005meteor} correlate poorly with human judgments. More recent metrics based on pre-trained neural networks, like BERTScore \cite{zhang2019bertscore} and BARTScore \cite{yuan2021bartscore}, achieve better semantic comprehension. However, they still struggle to align well with human standards in story evaluation. To address this, several works \cite{guan2020union,ghazarian2021MANPLTS,chen2023storyer,maimon-tsarfaty-2023-cohesentia} have conducted further training on story evaluation datasets or explored methods based on detailed analysis \cite{jiang2023tigerscore,xie2023deltascore} to improve performance. However, these explorations remain limited to \textit{short stories} generated from ROC and WP datasets. The \textit{criteria} used might also be restricted to predefined ones \cite{chhun2022hanna,xie2023can,wang2023perse}. These evaluation standards are inconsistent~\cite{yang2024makes}, and how well they align with actual readers' preferences remains unclear.
\paragraph{LLM-Based Evaluation.} The development of large language models also boosts LLM-based evaluations \citep{li2024leveraging,gao2024llm-based}. Through carefully designed prompts \cite{chen2023exploring,kim2023better} and helpful strategies \cite{chan2023chateval,saha2023branch,lee2024checkeval}, existing methods can achieve good correlation with humans. However, methods based on closed-source models can face problems of bias and inconsistency \cite{stureborg2024large}. Open-source LLM evaluators \cite{li2023autoj,kim2023prometheus}, on the other hand, include only a small portion of creative story evaluation data in their pretraining. Considering the important role of lengthy stories in people's daily lives, we attempt to explore how current LLMs handle lengthy story evaluation, compare different evaluation strategies, and propose a specialized evaluation model.

%% file: sections/3-benchmark.tex

\section{LongStoryEval Dataset}

\paragraph{Data Collection.} \label{sec:benchmark} 
Considering the high cost and time constraints of human annotations, we leverage large-scale online reviews from real readers. Our dataset comprises 600 newly published novels. Due to copyright restrictions, we release only plot and character summaries rather than the full book content. To ensure fairness in our experimental analysis, these books are verified to be absent from the pretraining dataset of our evaluated LLMs, avoiding data contamination issues\footnote{We also propose an anonymized test set for evaluating future LLMs, which can significantly mitigate data contamination concerns \cite{wang2023perse}. Details are in \S \ref{sec: contamination}.}~\cite{chang2023booookscore}. 
For each book, we collect its \textit{average rating score} and \textit{multiple reviews} from Goodreads, the largest book review platform. Each raw review consists of the reader's written critique along with a rating on a 1-5 scale. 

\paragraph{Review Processing.} As illustrated in Figure \ref{fig:review_process}, raw reviews are often unstructured and lack clarity. Prior works~\cite{gong2023coascore,lee2024checkeval} have shown that aspect-guided critiques enhance both readability and evaluation accuracy compared to direct scoring or relying on an unstructured overall review. Building on this insight, we reformat raw reviews by identifying user-mentioned aspects, extracting \textit{viewpoints} for each aspect, and summarizing these viewpoints into a concise \textit{overall assessment}. This process also involves refining the original language for clarity and brevity. The detailed processing prompt is provided in Table \ref{table:review_process}, with the temperature set to zero to prevent the introduction of new information. 

We first apply DeepSeek-v2.5 \cite{liu2024deepseek} to process the raw reviews. If a raw review is too ambiguous and the reformatted version has less than 40\% word overlap with the original text, we apply GPT-4o to process this raw review. If the overlap remains below the threshold after this second pass, the sample is filtered out. 

\begin{table*}[t]
\fontsize{6.2}{8.0}\selectfont
\begin{center}
\begin{tabular}{cccccccc}
\toprule
\textbf{Dataset} & \textbf{\# Stories} & \textbf{\# Samples} & \textbf{\# AVG Length} & \textbf{Review} & \textbf{Criteria} \\
\midrule
OpenMEVA \cite{guan2021openmeva}& 2,000 & 2.0K & 143 tokens & -  & PLOT\underline{(COH)}, \textbf{CHA}, WRI\underline{(FLU)} \\
HANNA \cite{chhun2022hanna} & 1,056  & 19.0K & 375 tokens &  -  & PLOT\underline{(COH,SUR)},  WOR\underline{(COM)},  \textbf{EMO(EMP)}, ENJ\underline{(ENG)}, \textbf{EXP(REL)}\\
StoryER-Rate \cite{chen2023storyer} & 12,669 & 45.9K & 493 tokens &Overall & PLOT\underline{(STR)}, CHA\underline{(CHAR)}, WRI\underline{(STY)}, EXP\underline{(GENRE)} \\
Xie \cite{xie2023can}& 200  & 1K &  79 tokens & - &   PLOT\underline{(COH)}, WRI\underline{(FLU)}, WOR\underline{(COMM)}, ENJ\underline{(INT)}, \textbf{EXP(REL)} \\
Per-DOC \cite{wang2023perse} & 596  & 8.9K & 2.5K tokens &  Overall & PLOT\underline{(ADAP,SUR,END)}, \textbf{CHA}, ENJ\underline{(INT)}\\
\midrule
\textbf{LongStoryEval} & 600  & 340K & 121K tokens  & Aspect-Guided & \textbf{PLOT, CHA, WRI, THE, WOR, EMO, ENJ, EXP}\\

\bottomrule
\end{tabular}
\end{center}
\caption{LongStoryEval and existing story evaluation datasets. ``{Criteria}'' denotes the considered aspects -- \textit{PLOT: plot \& structure, CHA: characters, WRI: writing \& language, THE: themes, WOR: world-building \& setting, EMO: emotional impact, ENJ: enjoyment \& engagement, EXP: expectation fulfillment}. Abbreviations of existing datasets' aspects are detailed in Table \ref{table:abbreviation}. Our criteria structure encompasses the previous inconsistent criteria, with overlapping top-level aspects shown in \textbf{bold} and covered sub-aspects \underline{underlined}. }

\label{table:eval_dataset} 
\end{table*}

\begin{figure}[t]
    \centering
        \includegraphics[width=.47\textwidth]{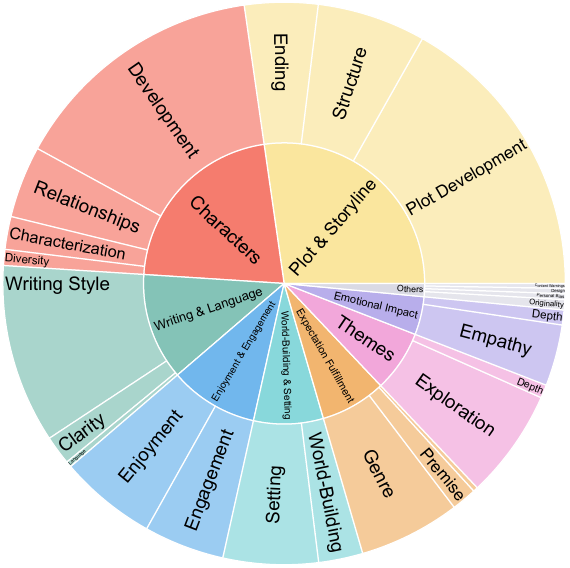}
        \caption{The distribution \protect\footnotemark of the evaluation aspects in readers' reviews.}
        \label{fig:aspect_distribute}
\end{figure}

\paragraph{Criteria Analysis.} Through our review process, we extract over 1000 user-mentioned aspects and analyze the most frequently referred ones. 
We organize these aspects into a hierarchical criteria structure, referring to existing evaluation works \cite{guan2021openmeva,chhun2022hanna} and literary studies \cite{halliwell1998aristotle, herman2011basic}. 
Specifically, we begin by analyzing the eight top-level aspects and use LLMs to identify potential sub-aspects. After further analysis and refinement, we establish our criteria structure (Table \ref{table:detailed_criteria}). Figure \ref{fig:aspect_distribute} shows the distributions of these aspects. Additional discussion can be found in \S \ref{appendix:criteria}. Among the top-level aspects, some focus on objective qualities of the novel (i.e., plot \& structure, characters, writing \& language, world-building \& setting, and themes), while others capture more subjective reader experiences (i.e., emotional impact, overall enjoyment \& engagement, and expectation fulfillment).

\paragraph{Review Organization.} After criteria analysis, we organize the 
 extracted viewpoints by grouping them under the same criteria as their corresponding \textit{critiques}. For example, as shown in Figure \ref{fig:data_construction}, the separate viewpoints for ``plot development'' and ``plot coherence'' are listed under ``Plot \& Structure'', forming the corresponding critique.

\begin{figure}[t]
    \centering
    \includegraphics[width=.99\linewidth]{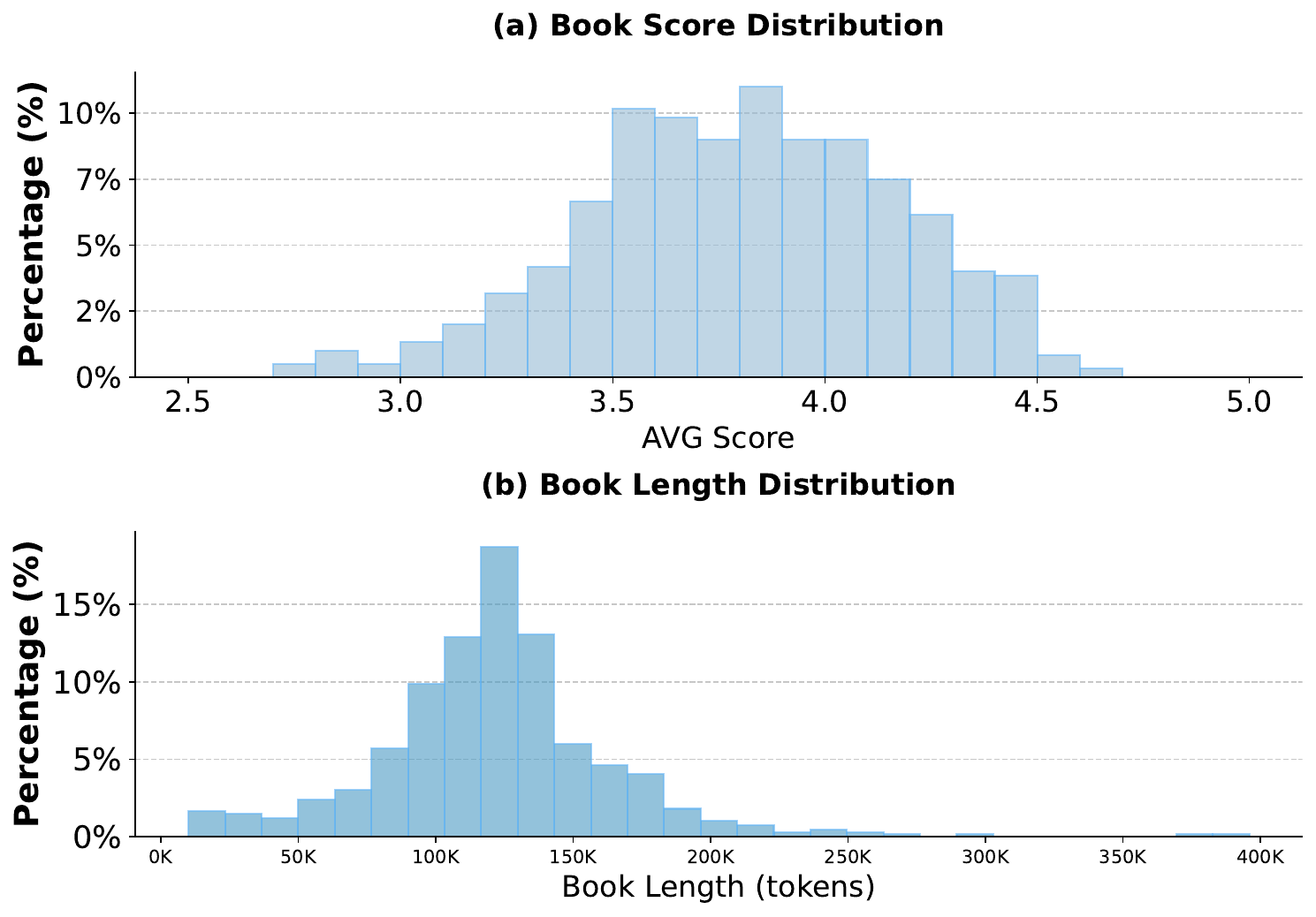}
    \caption{Average score distribution and book length distribution in LongStoryEval.}

    \label{fig:score_distribution}
\end{figure}

\paragraph{Statistics and Comparison.} Our benchmark dataset includes: (1) 600 newly published books with their metadata, including titles, genres, and premises; (2) An average rating score for each book, along with its rating distribution from 1-5 stars; (3) Multiple reviews for each book, organized as aspect-guided critiques, an overall assessment, and a final rating score; (4) Reviewer metadata, including rating score distribution and self-introduction (if available).

Compared to existing story evaluation benchmarks (Table \ref{table:eval_dataset}), \textbf{LongStoryEval} is the first to focus specifically on lengthy stories, with 
an average length of 121K tokens and a maximum of 397K. The length distribution is shown in Figure \ref{fig:score_distribution} (b). Unlike previous benchmarks, which rely on annotators to evaluate stories based on limited predefined criteria\footnote{While some benchmarks allow user-defined criteria, this is typically on a very small scale.}, we collect real-world reader reviews and derive evaluation criteria through systematic analysis. This data-driven approach ensures that the criteria structure better reflects actual reader standards. As shown in Table \ref{table:eval_dataset}, our evaluation criteria structure covers all key aspects identified in prior works. Additionally, our organized reviews demonstrate a multi-aspect-guided reasoning process for the evaluation score, enhancing interpretability and providing greater granularity for training story evaluation models.

\footnotetext{To avoid genre bias, this distribution includes equal books from each of the genres: Romance, Fantasy, Thriller, Mystery, Historical Fiction, Science Fiction, and Young Adult.}

%% file: sections/4-method.tex
\section{Method}
Given a book-length story consisting of several chapters $\{c_1, ..., c_n\}$ and an evaluation criteria list $\{a_1,...,a_m\}$, aspect-specific critique/review $r_i$ and  aspect-specific score $s_i$ will be generated for each $a_i$. All critiques will then be summarized into an overall assessment $R$, accompanied by an overall rating score $S$. 
In this work, we explore and compare three methods for evaluating lengthy stories (\S \ref{sec:lengthy_process}; Figure \ref{fig:framework}): aggregation-based, incremental-updated, and summary-based evaluations. We then propose a specialized model that uses the efficient summary-based strategy, as detailed in \S \ref{sec:novelcritique}.

\subsection{Lengthy Story Evaluation Methods} \label{sec:lengthy_process}

\begin{figure}[t]
\centering
    \includegraphics[width=.48\textwidth]{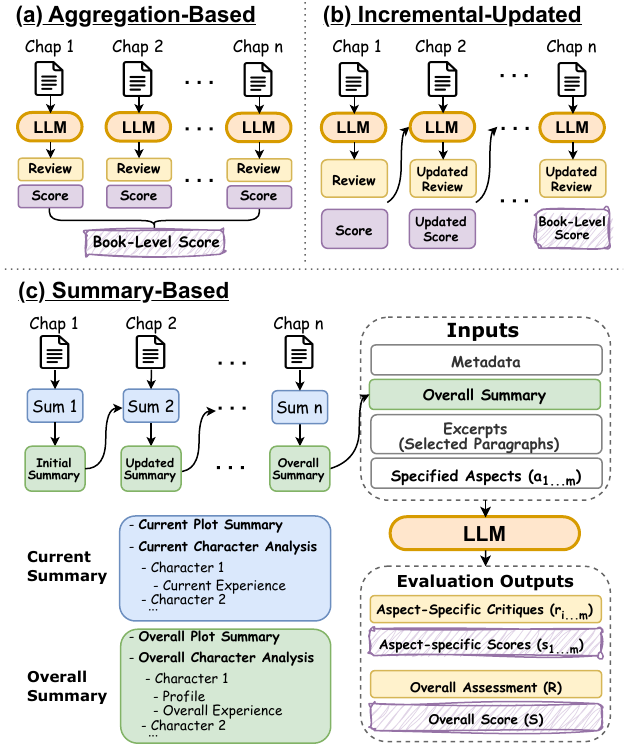}
            \caption{Overview of three evaluation methods (\S \ref{sec:lengthy_process}). Here we illustrate the complete inputs and outputs in the summary-based structure. The other two methods similarly incorporate metadata and specified aspects as inputs to generate aspect-specific and overall evaluations (reviews and scores). For these two types, each chapter's evaluation includes the current content and previous summaries as input, ensuring contextual awareness. Detailed prompt appears in Table \ref{table:prompt_eval}.}
            \label{fig:framework}
\end{figure}

\paragraph{Aggregation-Based Evaluation.} As illustrated in Figure \ref{fig:framework} (a), each chapter is evaluated individually, 
and the chapter-level scores are subsequently averaged as the book-level score. These chapter-level scores can refer to either aspect-specific scores or the overall score. For each chapter's evaluation, we provide the LLMs with the book's metadata, the current chapter, and a plot summary of previous chapters to ensure contextual awareness.
\vspace{-3px}
\paragraph{Incremental-Updated Evaluation.} This method assumes that a reader's opinion of a book evolves during the reading process. As illustrated in Figure \ref{fig:framework} (b), the model updates evaluations (both reviews and scores) progressively as it processes each chapter. At each step, the model receives the summary and evaluations from the previous chapters, processes the current chapter, and updates the reviews and scores. This process continues iteratively until the final chapter is reached. 

\vspace{-3px}
\paragraph{Summary-Based Evaluation.} \label{sec:summary_eval} 
A more intuitive approach involves reading the entire book first to form an overall impression before evaluation. A comprehensive overview of a lengthy story should include key aspects such as plot, characters, and writing style, similar to Wikipedia-style novel introductions. As illustrated in Figure \ref{fig:aspect_distribute}, these elements are also frequently mentioned by real readers. Therefore, we condense the story into: \textit{plot summary, character analysis}, and \textit{writing excerpts} (selected paragraphs to reflect the writing style). 
These elements can effectively capture additional aspects such as themes and overall enjoyment. 

Our summary is generated through incremental summarization, which aligns better with human preferences \cite{chang2023booookscore}. As shown in Figure \ref{fig:framework} (c), at each summarization step, we provide the current chapter and the previous summary (plot and character), generate a summary of the current chapter, and update the overall summary. Detailed explanations and prompts are displayed in \S \ref{appendix: summarization}.

\vspace{-3px}
\subsection{Proposed Model: \textit{NovelCrtique}} \label{sec:novelcritique}
As mentioned in \S \ref{sec:related_work} and confirmed by our experiments in \S \ref{sec:main_results}, closed-source methods, while outperforming open-source alternatives, still lack consistency and exhibit poor alignment with human evaluations. To address these limitations, we introduce \textit{NovelCritique}, a specialized model for evaluating long-form stories. NovelCritique follows the summary-based structure, which can achieve comparable results to aggregation-based evaluations (Table \ref{table:results}) while being much more efficient. The only deviation from the framework in Figure \ref{fig:framework} (c) is that our real-reader reviews lack aspect-specific scores. For each training sample with a criteria list $\{a_1,...,a_m\}$, the model outputs consist of: aspect-specific critiques, an overall assessment, and an overall score. 
During inference, when an aspect-specific score is needed, the model applies the generated critiques of this aspect, summarizes them into an overall assessment, and produces a score that becomes the aspect-specific score. 



\paragraph{Review Bias Mitigation.} We address the selection bias in providing reviews, as users who give moderate ratings are less likely to write reviews (\S \ref{sec:review_bias}). Training on all collected reviews would create bias in model predictions.
To counter this, we filter training reviews across all rating levels (1-5) to match each book's rating distribution. For instance, if a book has mostly 3-star ratings, but 5-star reviews are disproportionately overrepresented due to this bias, we filter out extra 5-star reviews based on the book's rating distribution. 
\paragraph{Rating Score Normalization\protect\footnotemark.} 
Users' rating standards vary significantly, with some being strict and others more moderate. To normalize the ratings \cite{ShalabiS2006Normalization}, we adjust each rating score $S$ as follows:

\begin{equation} 
   S' = \frac{S-\mu_u}{\sigma_u}\times \sigma_{\text{plat}} + \mu_{\text{plat}},
\end{equation}
where $\mu_u, \sigma_u$ denote the mean and standard deviation of the current user's rating distribution, and $\mu_{\text{plat}}, \sigma_{\text{plat}}$ represent platform-wide statistics. 

\footnotetext{This normalization is applied only to training samples, while for evaluation, we use the average of all original ratings.}
\paragraph{Training.} 
We train NovelCritique via instruction tuning \cite{ouyang2022instruction} using cross-entropy loss. The instruction details are provided in Table \ref{table:instrction_prompt}. We select Llama 3.1-8B \cite{dubey2024llama} as the base model due to its strong performance among open-source alternatives. The training loss of each sample is calculated as:
\begin{equation}\footnotesize
    -\log P(r_{i\le m}, R, S'|X_{\text{Instruct}, 
 \text{Metadata}, \text{Sum}, \text{Exceprts}},a_{i\le m}).
\end{equation}

%% file: sections/5-experiments.tex
\vspace{-4px}
\section{Experiments}

We conduct experiments using the LongStoryEval dataset. From the 600 books, we designate 150 books as the test set (Tables \ref{tab:books_info}-\ref{tab:books_info_2}), ensuring diversity across genres and score distributions. 
\subsection{Training Setup of NovelCritique} 
Our training set consists of the remaining 450 books and 176K filtered reviews (after mitigating review bias). Input summaries are generated via incremental summarization (\S \ref{appendix: summarization}) using the GPT-4o model. We finetune the base Llama 3.1-8B model for three epochs with a learning rate of $1e^{-5}$ and a batch size of 32. The LoRA parameters \cite{Hu2022LoRA} are configured as $r=64$ and $alpha=16$. The training was conducted on four A6000 GPUs, taking approximately $125$ hours.  

\subsection{Baselines}\label{sec:baselines}
\paragraph{LLM-Based Lengthy Story Evaluation.} We conduct experiments to compare the effectiveness of three evaluation methods (detailed in \S \ref{sec:lengthy_process}): aggregation-based, incremental-updated, and summary-based evaluations. Building on research showing that LLM-based evaluation benefits from detailed criteria definitions \cite{Chhun2024do}, we establish evaluation criteria for eight top-level aspects based on literature standards, as detailed in \S \ref{appendix:criteria_def}. The evaluation prompts are provided in Table \ref{table:prompt_eval}. 

\paragraph{Backbone LLMs.} We experiment with five backbone models: GPT-4o, GPT-4o-mini, DeepSeek-v2.5 \cite{liu2024deepseek}, Mixtral 8×7B-Instruct \cite{jiang2024mixtral}, Llama 3.1-70B-Instruct, and Llama 3.1-8B-Instruct \cite{dubey2024llama}. As shown in Table \ref{tab:knowledge_cut}, all these models were trained on data predating year \textit{2024} to ensure fairness in evaluation. To improve stability, we apply greedy decoding for open-source models and set the temperature to zero for closed-source models. Since closed-source LLMs still produce variations (they do not use pure greedy decoding), we report the average rating score across five generations.





\begin{table*}[t]
\fontsize{8}{10.3}\selectfont
\begin{center}
\begin{tabular}
{ll|rrrrrrrr|r}
 \toprule
 &  & \multicolumn{1}{c}{\textbf{PLOT}} & \multicolumn{1}{c}{\textbf{CHA}} & \multicolumn{1}{c}{\textbf{WRI}} & \multicolumn{1}{c}{\textbf{WOR}} & \multicolumn{1}{c}{\textbf{THE}} & \multicolumn{1}{c}{\textbf{EMO}} & \multicolumn{1}{c}{\textbf{ENJ}} & \multicolumn{1}{c}{\textbf{EXP}} & \multicolumn{1}{|c}{\textbf{Overall}}  \\
\midrule
\multirow{2}{*}{\textbf{\begin{tabular}[c]{@{}l@{}}One-Pass\\ 
(Subset)\end{tabular}}} & GPT-4o &3.3&4.1&7.9&0.8&3.3&-1.2&-3.2&8.4&5.5\\
& DeepSeek-v2.5 & 4.4&3.5&4.8&-0.9&3.3&-1.1&-1.3&9.4&4.8\\
\midrule
\multirow{6}{*}{\textbf{\begin{tabular}[c]{@{}l@{}}Aggregation\\ 
-Based\end{tabular}}}  & GPT-4o &14.3&16.7&10.2&7.9&10.4&9.7&9.1&14.1&15.2\\
 & DeepSeek-v2.5 &17.2&15.8&7.0&7.1&11.0&14.2&11.1&16.7&15.1\\
 & GPT-4o-mini &14.2&17.2&7.2&4.4&9.5&8.9&8.1&15.1&12.3\\
 & Llama 3.1-70B &19.6&13.8&2.3&13.8&13.4&7.7&11.5&18.9&13.8\\
 & Llama 3.1-8B &15.5&8.5&-1.4&2.8&12.3&7.5&7.0&13.7&11.6\\
 & Mixtral 8$\times$7B &9.5&4.0&2.5&-0.2&8.9&9.5&10.2&6.8&9.0\\
 \midrule
\multirow{6}{*}{\textbf{\begin{tabular}[c]{@{}l@{}}Incremental\\ 
-Updated\end{tabular}}} & GPT-4o &8.0&9.1&9.1&11.7&10.5&12.3&12.1&11.5&10.9\\
 & DeepSeek-v2.5 &8.9&12.2&9.0&8.6&12.5&12.3&6.6&12.2&11.6\\
 & GPT-4o-mini &7.9&10.8&6.7&7.4&8.5&11.6&8.5&10.7&9.3\\
 & Llama 3.1-70B &9.3&13.3&4.1&1.7&8.7&4.9&4.6&6.1&9.9\\
  & Llama 3.1-8B & 7.0&7.1&4.4&2.5&1.9&8.0&7.8&5.1&6.7\\
 & Mixtral 8$\times$7B &4.2&10.8&4.4&6.6&5.8&2.3&5.8&2.6&4.2\\
 \midrule
\multirow{7}{*}{\textbf{\begin{tabular}[c]{@{}l@{}}Summary\\ 
-Based\end{tabular}}} & GPT-4o &15.3&17.8&4.5&5.0&7.2&12.6&11.8&14.0&13.4\\
 & DeepSeek-v2.5 &13.4&12.2&1.8&-3.8&7.1&8.9&13.2&15.1&14.4\\
 & GPT-4o-mini &8.7&7.5&5.4&4.8&11.1&11.6&8.3&7.9&9.7\\
 & Llama 3.1-70B &11.2&10.8&-1.6&5.3&12.4&9.2&11.4&14.5&13.0\\
 & Llama 3.1-8B &10.4&14.1&4.9&9.1&9.6&15.3&14.5&12.3&12.4\\
 & Mixtral 8$\times$7B &7.8&7.4&7.1&-0.5&-4.0&5.6&9.4&6.7&8.3\\
& \textbf{NovelCritique-8B } &\textbf{27.1}&\textbf{27.0}&\textbf{24.1}&\textbf{18.3}&\textbf{24.3}&\textbf{27.8}&\textbf{21.1}&\textbf{25.5}&\textbf{27.7}\\
\bottomrule
\end{tabular}
\end{center}
\caption{The system-level Kendall correlations between the human-assigned scores and model-generated evaluations. We report the correlation between aspect-specific scores and the overall score.}
\label{table:results} 
\end{table*}

%% file: sections/6-results.tex

\subsection{Main Results} \label{sec:main_results}
Following prior evaluation works, we use Kendall-Tau correlations \cite{kendall1938new} to measure agreement between human evaluations and model-predicted scores. 
For each evaluation method, we generate aspect-specific scores and the overall scores, then compute their correlation with human-assigned ratings (i.e., the average rating of each book). As shown in Table \ref{table:results}, \textbf{NovelCritique} demonstrates the highest correlation with human ratings across both overall scores and multiple evaluation aspects. This demonstrates the effectiveness of training on domain-specific story review data.

The primary issue with closed-source LLMs is their inconsistency.  
Even with temperature=0 and low top-p settings, the results exhibit significant variability. This inconsistency is likely due to the long context windows, which increase the likelihood for models to focus on uncertain or less relevant story elements. 
Notably, this inconsistency issue is less pronounced in short story evaluation tasks \cite{Chhun2024do}.
To mitigate this, we average scores across five evaluation runs. While this improves stability, it also significantly increases computational overhead. \textbf{The cost is particularly high for incremental-updated and aggregation-based methods} as they require processing the entire book context for each evaluation run. 

\subsection{Analysis}\label{sec:analysis}

\paragraph{Which evaluation aspects mostly affect the final rating?}\footnote{The analysis is based only on experiments using our test set; further exploration is needed to reach more robust conclusions.}
As the results shown in Table \ref{table:results}, for objective aspects, \textit{plot} and \textit{characters} are the most influential. \textit{Themes} are important but appear to be secondary considerations for most readers. \textit{World-building} and \textit{writing quality} are the least influential aspects, likely because most stories show similar levels in these areas (except for particularly exceptional books). For subjective aspects, \textit{emotional impact}, \textit{enjoyment \& engagement}, and \textit{expectation fulfillment} all play critical roles. 



\begin{figure*}[t]
    \centering
    \includegraphics[width=0.99\linewidth]{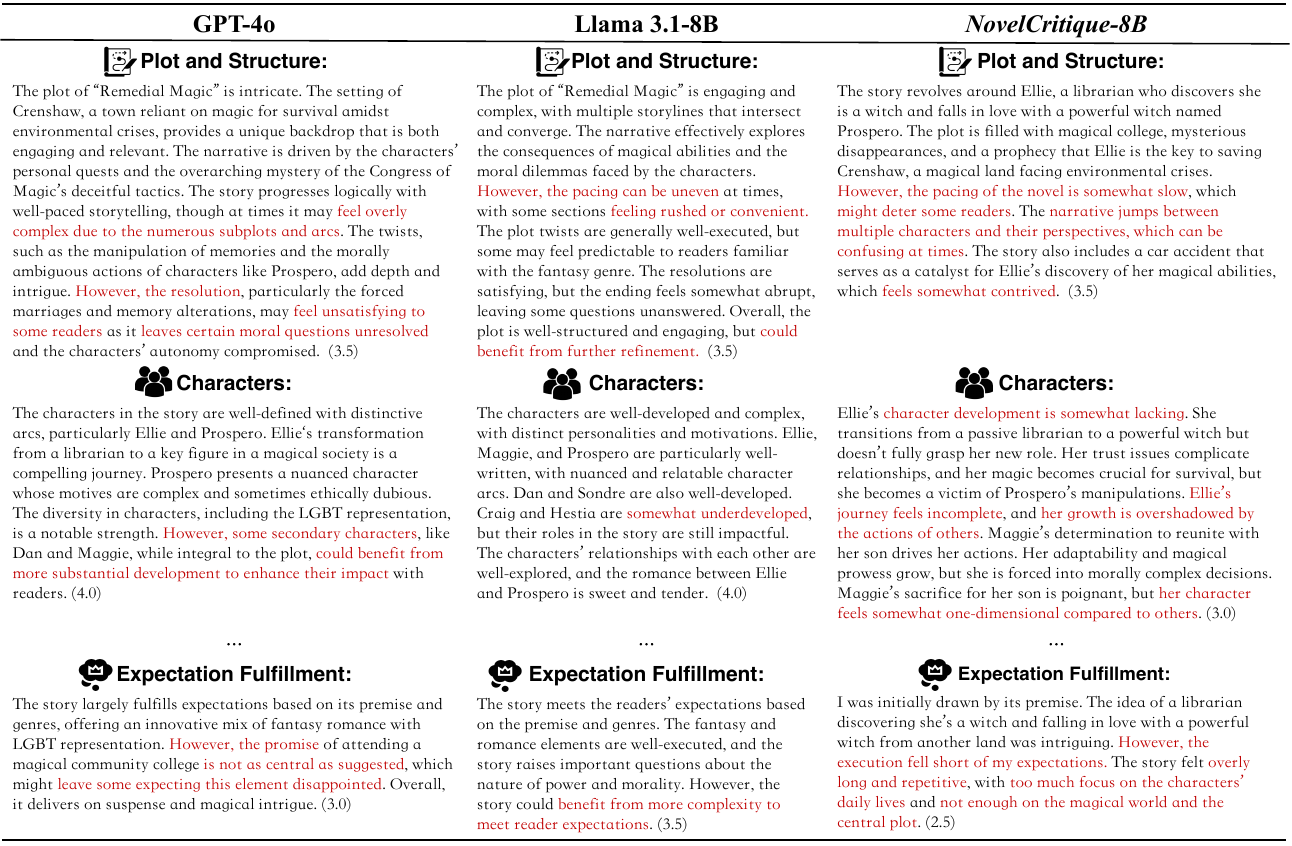}
    \caption{Critiques for ``Remedial Magic'' (AVG human rating: 2.9). The generated weaknesses are colored in red.}
    \label{fig:cases}
\end{figure*}
 
 \begin{figure}[t]
    \centering
    \includegraphics[width=.47\textwidth]{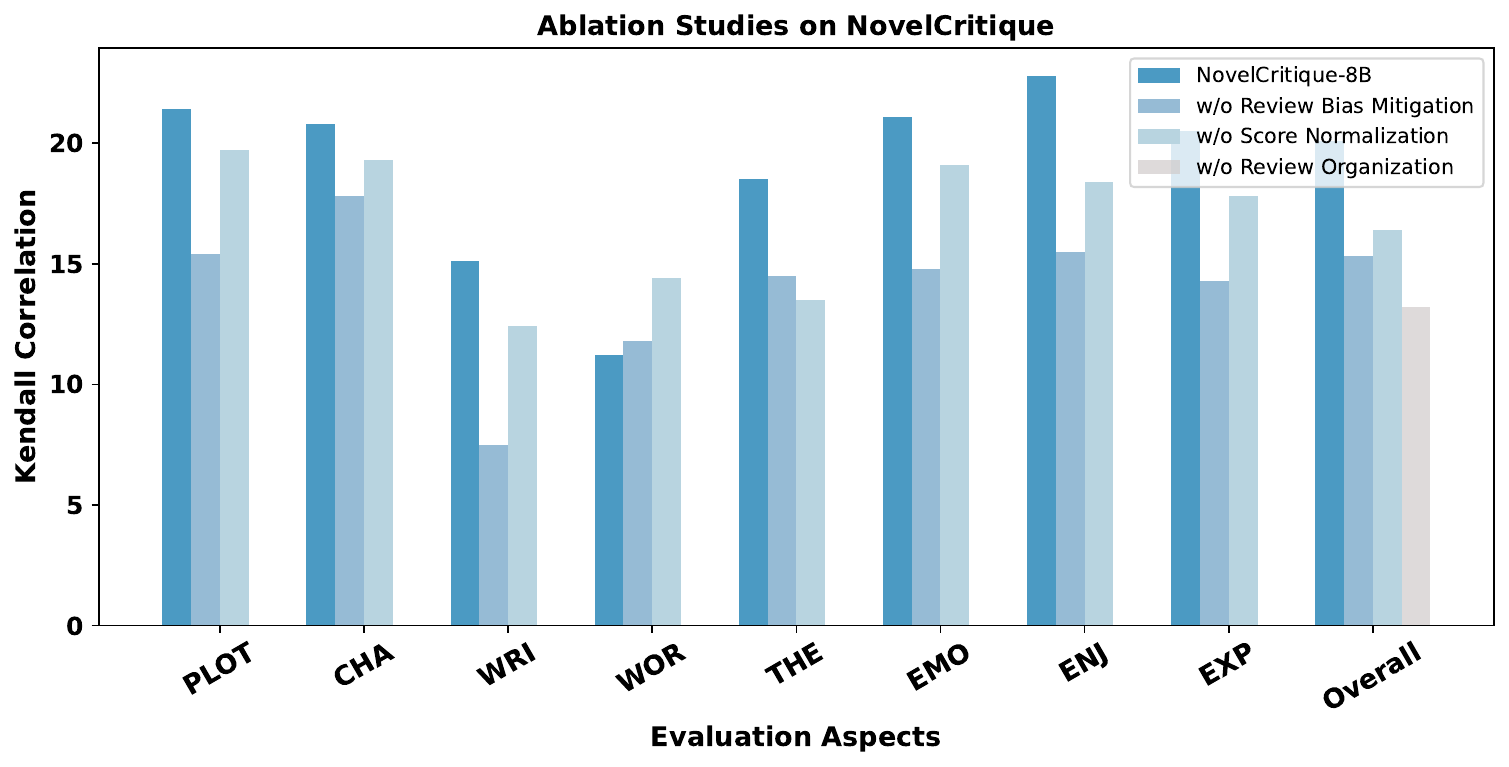}
    \caption{Ablation studies on NovelCritique.}

    \label{fig:ablation}
\end{figure}

\paragraph{Which long story evaluation strategy is more effective?} Prior to comparing our proposed methods, we assess whether existing models can effectively evaluate entire books in a single pass. We test a subset of books within the 128K token context. The one-pass results (lines 1-2 in Table \ref{table:results}) reveal a poor correlation with human ratings. Even when prompted to generate summaries first, these models often produce generic critiques that fail to capture the nuances of specific stories. 

Regarding the three methods discussed in \S \ref{sec:lengthy_process}, aggregation-based and summary-based approaches demonstrate superior performance. While the incremental-updated method seems promising in theory, it faces two key limitations. First, it requires additional instructions, forcing models to both comprehend the current segment and consider its impact on previous evaluations. This problem is especially serious for less powerful models like Llama 3.1-8B and Mixtral 8$\times$7B. Second, it suffers from inconsistency that accumulates over multiple updates. Given these constraints, we recommend using aggregation-based and summary-based methods until significant improvements in LLM capabilities emerge. 

The main advantage of aggregation-based methods is their ability to access all details of a long story, enabling more thorough measurement and scoring. Summary-based models, on the other hand, offer two main benefits: (1) \textbf{Efficiency}—they require less time and resources per evaluation. By generating a single high-quality summary, we can reuse it across multiple evaluations, leading to more stable and robust results. (2) \textbf{Potential for early evaluation}—before completing the entire story, authors can receive reviews and scores based on their plot structure, character design, and writing samples. These early assessments can closely estimate judgments of their finished work. 


\paragraph{Do detailed summaries improve summary-based evaluation?} 
As displayed in Table \ref{tab: summ}, replacing the overall summary with detailed chapter-guided summaries leads to a slight increase. We suggest that a more detailed summary can better reflect a story's quality. However, longer summaries require more memory and involve more complex reasoning. It is important to find a balance between the level of detail and length, which could be explored in future works.

\paragraph{Is high-quality summaries necessary for summary-based evaluation?}
 To assess the importance of summary quality, we replace GPT-4o-generated summaries with those produced by GPT-4o-mini (around 0.03\% prices of GPT-4o). The results (Table \ref{tab: summ}) show no significant decline in performance. This suggests a cost-efficient approach: generating summaries with GPT-4o-mini and then conducting evaluations using more advanced models like DeepSeek-v2.5 or NovelCritique.

\paragraph{Ablation Studies on NovelCritique.}
We verify the effectiveness of our designs in NovelCritique, including raw review organization, review bias mitigation, and rating score normalization. The results in Figure \ref{fig:ablation} demonstrate their effectiveness.


\subsection{Qualitative Results}
In Figure \ref{fig:cases}, we present evaluation results from different models. We find that many models tend to focus more on the story's strengths, offering only limited commentary on its weaknesses. This tendency will also lead existing models to assign good scores for stories that humans consider poor. While we have tried to address this by asking models to provide advantages and disadvantages, this approach causes models to become excessively critical. Current models still struggle to generate nuanced critiques that closely align with human preferences, particularly in reflecting detailed evaluations. Critiques for a well-written story are displayed in \S \ref{sec:additional_cases}. 

%% file: sections/7-conclusion.tex
\section{Conclusion} 

This work explores the underexplored problem of evaluating book-length stories, addressing three core questions: 
(1) What evaluation aspects matter most to real readers? (2) What are the most effective methods for evaluating lengthy stories? (3) What challenges arise in LLM-based evaluation and how can they be addressed? To tackle these questions, we introduce \textit{LongStoryEval},  a large-scale dataset comprising average rating scores and well-formatted reviews. Through analysis of these reviews, we propose a \textit{criteria structure} that reflects human standards. Our experiments reveal the critical aspects influencing final ratings, and demonstrate the effectiveness of aggregation- and summary-based evaluations. While aggregation-based methods provide detailed and comprehensive evaluations, summary-based methods excel in efficiency and offer potential for early-stage evaluations. 
Acknowledging the limitations of existing LLMs, such as inconsistency and imperfect alignment with human preferences, we propose \textit{NovelCritique}, an 8B model that exhibits improved correlation with human evaluations.  
We hope this work inspires further research into evaluating lengthy stories and fosters advancements in both lengthy story evaluation and generation.

\section*{Limitations}
This work employs critiques and score generation for evaluation, which can be prone to inconsistencies. To mitigate this, we average results across multiple runs. Future work could explore alternative mitigation methods, such as employing pair-wise comparison instead of direct scoring. Comparisons often yield more stable results but come with higher computational costs, underscoring the need for more efficient comparison strategies. Sampling-based approaches \cite{xu2024data} also present a promising direction for generating more reliable scores. 

Our current evaluation emphasizes general assessment over personalized preferences. However, since our dataset contains anonymized reviewer information, future studies could explore personalized evaluation approaches tailored to individual tastes and reading habits.

\section*{Ethical Problems}
We acknowledge and strictly adhere to the Code of Ethics and Professional Conduct throughout this research. The potential ethical concerns are addressed as follows:
\paragraph{Data Source.} Our review data comes from publicly available content on the Goodreads website \footnote{\url{https://www.goodreads.com}}, which is accessible to anyone. Following previous works \cite{wan2019Fine,wan2018item}, we anonymize user IDs and review IDs to protect personal information. To mitigate the potential dissemination of harmful content in the raw reviews, we will only release our processed versions of reviews. For the books in our dataset, we collect all metadata from public Goodreads content and purchase electronic copies of the books. Considering copyright issues, only the book-level summaries will be released \cite{chang2023booookscore}, while full content remains accessible through publicly available titles and author information (Tables \ref{tab:books_info}-\ref{tab:books_info_2}). 
\paragraph{Copyrights.} As discussed before, we will only release processed versions to avoid potential ethical and copyright issues. Our data collection from public resources is for academic use only. To prevent commercial use, we will release our dataset under highly restrictive permissions that limit its use exclusively to academic research.

%% file: sections/appendix.tex
\section{Evaluation Criteria} \label{appendix:criteria}

\subsection{Human Criteria}
Table \ref{table:detailed_criteria} presents our detailed criteria structure with raw aspects mapped to standardized categories. Below, we detail our analyzed criteria structure, including the aspect names, definitions, and their corresponding references:
\paragraph{Plot and Structure:}~{}
\vspace{4px}

\noindent \textbullet\ \   \textbf{Plot Development}: Assess the progression of events within a story through their pacing, twists, conflicts, and resolutions, as mentioned in \citet{bell2004plot,forster1927aspects,brooks1992reading}.
\vspace{4px}

\noindent \textbullet\ \  \textbf{Structure}: Assess the organization of events through exposition, rising action, climax, falling action, and resolution, examining their coherence and logical flow, as mentioned in \citet{freytag1895technique,genette1980narrative}.
\vspace{4px}

\noindent \textbullet\ \  \textbf{Ending}: How well the ending reflects the consequences of the story’s main events and themes. A well-crafted ending provides closure or shows attractive potential, as mentioned in \citet{rimmon2003narrative,phelan1996narrative}.

\paragraph{Characters:}~{}
\vspace{4px}

\noindent \textbullet\ \  \textbf{Character Development}: Examine how well characters develop throughout a story by examining their backstories, motivations, and character growth arcs, as mentioned in \citet{mckee1997substance,james1884art,campbell2008hero}.

\vspace{4px}
\noindent \textbullet\ \  \textbf{Characterization}: Assess the representation of the characters, ensuring appeal, realism, and reliability, as mentioned in \citet{wood2008fiction,burroway2019writing}.

\vspace{4px}  \noindent \textbullet\ \  \textbf{Relationships}: Assess the interactions and chemistry between the characters, examining how their relationships drive the plot forward and engage readers, as mentioned in \citet{james1884art,chatman1980story}.

\vspace{4px}  \noindent \textbullet\ \  \textbf{Diversity}: How effectively the story portrays a diverse cast of characters, including not only the protagonist but also secondary and side characters, mentioned in \citet{hogan2003mind,adichie2009danger}.

\paragraph{Writing and Language:}~{}

\vspace{4px}  \noindent \textbullet\ \  \textbf{Writing Style}: Whether the author's choice of words, sentence structure, and paragraph organization conveys meaning effectively and creates an engaging atmosphere for readers, as mentioned in \citet{leech2007style,skorupski1976symbol}.
\vspace{4px}  

\noindent \textbullet\ \  \textbf{Language}: Assess the quality of vocabulary and syntax, focusing on issues like grammar and fluency, mentioned in \citet{pinker2003language, elbow1998writing}.

\vspace{4px}  
\noindent \textbullet\ \  \textbf{Readability}: Assess the clarity and readability of the entire story, as mentioned in \citet{burroway2019writing,king2000writing}.

\paragraph{Themes:}~{}

\vspace{4px}  

\noindent \textbullet\ \  \textbf{Exploration}: How well the author explores clear topics and themes through the narrative, as mentioned in \citet{forster1927aspects,booth1983rhetoric}.
\vspace{4px}  

\noindent \textbullet\ \  \textbf{Depth}: Measure the depth of the themes, such as their educational value and social commentary, as mentioned in \citet{booker2004seven,jameson2013political}.

\paragraph{World-Building and Setting:}~{}

\vspace{4px}  

\noindent \textbullet\ \  \textbf{World-Building}: Assess the realism and authenticity of the built world by examining its consistency and descriptive detail, as mentioned in \citet{roine2016imaginative,james1884art,coulton2017design}.
\vspace{4px}  

\noindent \textbullet\ \  \textbf{Setting}: Evaluate the accuracy of historical, cultural, geographic, and technical settings, as discussed in \citet{de2009historical, james1884art}. This might be more considered in historical stories.

\paragraph{Emotional Impact:}~{}

\vspace{4px}  

\noindent \textbullet\ \  \textbf{Empathy}: Evaluate how well the stories evoke emotional connections between readers and characters, as mentioned in \citet{mccabe1984makes,keen2007empathy,hogan2003mind}.
\vspace{4px}  

\noindent \textbullet\ \  \textbf{Depth}: Evaluate the stories' ability to create deep emotional connections, as discussed in \citet{ahmed2013cultural,keen2007empathy}.

\paragraph{Enjoyment and Engagement:}~{}

\vspace{4px}  

\noindent \textbullet\ \  \textbf{Enjoyment}: Evaluate how well a story creates enjoyment and interests, and builds anticipation for upcoming events, as mentioned in \citet{nell1988lost,zunshine2006we}.
\vspace{4px}  

\noindent \textbullet\ \  \textbf{Engagement}: Evaluate how well the story engages the readers through the reading process, as mentioned in \citet{nell1988lost,zunshine2006we}.

\paragraph{Expectation Fulfillment (Or Relevance):}~{}

\vspace{4px}

\noindent \textbullet\ \ \textbf{Genre}: How well the story fulfills genre expectations for elements like romance, suspense, and other genre-specific features. As mentioned in \citet{cawelt2014adventure,phelan1996narrative}.
\vspace{4px}

\noindent \textbullet\ \ \textbf{Premise}: How well the story fulfills expectations based on the premise, as mentioned in \citet{forster1927aspects,field2005screenplay}.

\paragraph{Others:}~{}

\vspace{4px}

\noindent \textbullet\ \  \textbf{Originality (Or Creativity)}: Assess the story's uniqueness and imaginative ideas, as mentioned in \citet{csikszentmihalyi1997flow}.
\vspace{4px}

\noindent \textbullet\ \  \textbf{Content Warnings}: Whether the story contains rude, unreasonable, or disrespectful elements, as mentioned in \citet{wyatt2016ethics}.
\vspace{4px}

\noindent \textbullet\ \  \textbf{Designment}: Assess the design of the book cover \cite{matthews2007judging}.
\vspace{4px}

\noindent \textbullet\ \  \textbf{Personal Bias}: Assess the feelings influenced by very subjective factors, such as individual preferences and personal experiences \cite{wang2023perse}.

\subsection{Condensed Definitions}\label{appendix:criteria_def}
We condense the definition of the top-level aspects as follows, which can be applied in our evaluation prompts:
\begin{enumerate}
\item \textbf{Plot and Structure (PLOT)}: Evaluate the plot development by examining pace, twists, conflicts, and their resolutions. Evaluate the story structure for coherence, logic, and complexity, paying attention to key elements like climax and ending.
\item \textbf{Characters (CHA)}: Evaluate how well the characters are drawn, considering their development, characterization (including realism, appeal, and relatability), relationships, and diversity.
\item \textbf{Writing and Language (WRI)}: Evaluate the writing style's ability to engage and captivate readers. Assess language quality by examining descriptions and dialogue. Measure the story's clarity and readability.
\item \textbf{World-Building and Setting (WOR)}: Evaluate the world-building and setting by assessing how detailed and well-described they are. Consider their authenticity, accuracy, or realism.
\item  \textbf{Themes (THE)}: Evaluate how well the themes are explored throughout the story and assess their depth.
\item \textbf{Emotional Impact (EMO)}: Evaluate the story's ability to evoke strong and deep emotional impact.
\item \textbf{Enjoyment and Engagement (ENJ)}: Evaluate how engaging and enjoyable the story is for readers.
\item \textbf{Expectation Fulfillment (EXP\footnote{This can also be considered as the relevance to the "Inputs"(premise and genres).})}: Evaluate how effectively the story meets the readers' expectations based on the premise and genres.

\end{enumerate}

\subsection{Criteria of Existing Benchmarks}
The considered aspects of existing benchmarks and their abbreviation are listed in Table \ref{table:abbreviation}.

\section{Dataset: LongStoryeval}
\subsection{Review Processing}
\begin{figure*}[t]
    \centering
    \includegraphics[width=\textwidth]{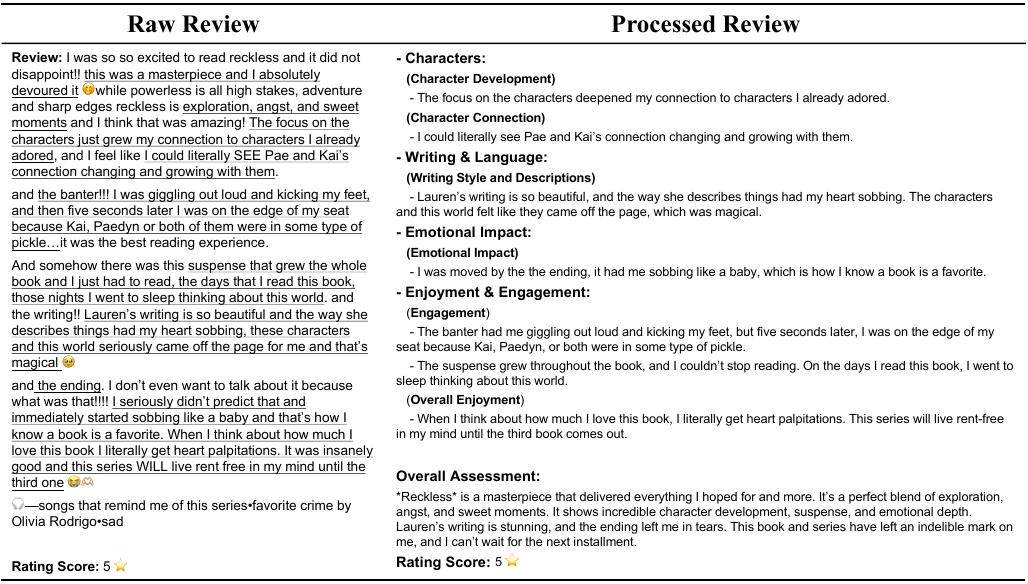}
    \caption{Example of raw review processing. We identify user-mentioned aspects (shown in brackets), then extract viewpoints (underlined texts) for each aspect, and summarize these viewpoints into an overall assessment. These viewpoints will be organized based on the criteria structure shown in Table \ref{table:detailed_criteria}.}
    \label{fig:review_process}
\end{figure*}

Figure \ref{fig:review_process} displays an example of our raw review processing, while the processing prompt is detailed in Table \ref{table:review_process}. Our dataset sampling shows high accuracy (96\%), which is sufficient for our model training. Our experimental improvements demonstrate its quality, though further human refinement could be considered in the future. 

\subsection{Statistics of Review Bias} \label{sec:review_bias}
Review bias refers to the selection bias in providing reviews, where users who give moderate ratings are less likely to leave reviews. We evaluate the average ratio of review numbers among ratings for each rating scale. The statistics show: 5 stars (22\%), 4 stars (19\%), 3 stars (17\%), 2 stars (22\%), and 1 star (31\%).

\begin{table}[t]
\fontsize{7}{9}\selectfont
\centering
\begin{tabularx}{.5\textwidth}{lX}
\toprule
\textbf{Dataset} & \textbf{Aspects} \\
\midrule
\multirow{3}{*}{\begin{tabular}[c]{@{}l@{}}OpenMEVA\\ 
\cite{guan2021openmeva}\end{tabular}} & COH: coherence [consistency, causal and temporal relationship], CHA: character behavior, FLU: fluency [semantic repetition, paraphrases] \\
\midrule
 \multirow{2}{*}{\begin{tabular}[c]{@{}l@{}}HANNA\\ 
\cite{chhun2022hanna}\end{tabular}}& COH: coherence, SUR: surprise, COM: complexity, EMP: empathy, ENG: engagement, REL: relevance.\\
 \midrule
 \multirow{3}{*}{\begin{tabular}[c]{@{}l@{}}StoryER-Rate\\ 
\cite{chen2023storyer}\end{tabular}} & STR: structure, CHAR: characterization [character shaping], STY: writing style, GENRE: genre expectations. \\
 \midrule
 \multirow{2}{*}{\begin{tabular}[c]{@{}l@{}}Xie\\ 
\cite{xie2023can}\end{tabular}} & FLU: fluency, COH: coherence, COMM: commonsense, INT: interestingness, REL: relevance. \\
 \midrule
 \multirow{2}{*}{\begin{tabular}[c]{@{}l@{}}Per-DOC\\ 
\cite{wang2023perse}\end{tabular}}& ADAP: plot adaptability, SUR: surprise, END: ending, CHA: characters, INT: interestingness.\\
 \bottomrule
\end{tabularx}
\caption{Existing story evaluation benchmarks and their considered aspects.}
\label{table:abbreviation}
\end{table}

\subsection{Contamination Issues} \label{sec: contamination}
Contamination problems mean the datasets are exposed to LLMs. For contaminated cases, LLMs might show unfairly high performance \cite{chang2023booookscore,wang2023perse}. 
However, anonymization and summarization can largely mitigate these issues \cite{wang2023perse}. Since our dataset is already summarized, we conduct additional anonymization by replacing identifiable elements such as character names and locations. This anonymized test set helps ensure fair evaluation of future LLMs. Additionally, we will release our processing code to facilitate the collection of newly published books.

\subsection{Approximate Exploration of Significant Aspects} \label{sec:appro}
Since individual aspect ratings are unavailable in the ground truth data, we attempt to approximate aspect-specific ratings by calculating the average of ratings from these reviews covering each specific aspect. As the results shown in Table \ref{tab:appro}, plot, characters, and overall enjoyment show clear significance. However, this approximation is imperfect, as the scores are influenced by multiple mentioned aspects rather than representing a single aspect. Therefore, these findings should be considered as a reference, with the results in Table \ref{table:results} offering more reliable evidence.

\begin{table}[t]
\fontsize{6.5}{7}\selectfont
\centering
\begin{tabular}{ccccccccc}
\toprule
\textbf{PLOT} & \textbf{CHA} & \textbf{WRI} & \textbf{WOR} & \textbf{THE} & \textbf{EMO} & \textbf{ENJ} & \textbf{EXP} \\ \midrule
79.9 & 83.2 & 72.0 & 66.4 & 72.3 & 74.0 & 79.1 & 72.4 \\ \bottomrule
\end{tabular}
\caption{Approximate correlation scores (\S \ref{sec:appro}).}
\label{tab:appro}
\end{table}

\section{Summarization Method}\label{appendix: summarization}

As mentioned in Section \ref{sec:summary_eval}, in our summary-based methods, we apply the summaries generated through an incremental-update process. Specifically, we update both plot and character summarizations. The detailed prompts are shown in Table \ref{table:prompt_summary_begin} and \ref{table:prompt_summary}. For characters, we summarize their profiles and overall experiences. For plots, considering that non-linear narrative structures can affect readers' reading experience, we instruct the model to maintain the original plot structure.

 \section{Experimental Details}\label{appendix:other_results}
 
\subsection{Knowledge Cutoff of Baseline Models.}
As shown in Table \ref{tab:knowledge_cut}, all of the baseline models are trained with data before 2024, not containing our dataset.

\begin{table}[h]
\fontsize{9}{10}\selectfont
\centering
\begin{tabular}{lcc}
\toprule
\textbf{Models} & \textbf{Knowledge Cutoff}  \\
\midrule
GPT-4o & October, 2023\\
GPT-4o-mini & October, 2023\\
DeepSeek-v2.5 \cite{liu2024deepseek} & November, 2023\\
Mixtral 8$\times$7B \cite{jiang2024mixtral} & December, 2023\\
Llama 3.1-8B \cite{dubey2024llama} & December, 2023\\
 \bottomrule
\end{tabular}
\caption{Knowledge cutoff of baseline models.}
\label{tab:knowledge_cut}
\end{table}

\subsection{Additional Discussions}
We find that during the incremental evaluation, if the LLMs generate positive opinions for the earlier paragraphs, then they are likely to maintain this tendentiousness. Even if there are problems in the latter chapters, they will not significantly change their assessment. The same pattern occurs in reverse. 


\begin{table*}[t]
\fontsize{9}{11}\selectfont
\begin{center}
\begin{tabular}
{ll|rrrrrrrr|c}
 \toprule
 &  & \multicolumn{1}{c}{\textbf{PLOT}} & \multicolumn{1}{c}{\textbf{CHA}} & \multicolumn{1}{c}{\textbf{WRI}} & \multicolumn{1}{c}{\textbf{WOR}} & \multicolumn{1}{c}{\textbf{THE}} & \multicolumn{1}{c}{\textbf{EMO}} & \multicolumn{1}{c}{\textbf{ENJ}} & \multicolumn{1}{c}{\textbf{EXP}} & \multicolumn{1}{|c}{\textbf{Overall}}  \\
\midrule
\multirow{2}{*}{\textbf{\begin{tabular}[c]{@{}l@{}}GPT-4o\end{tabular}}}  &  w/ definitions&15.3&17.8&4.5&5.0&7.2&12.6&11.8&14.0&13.4\\
 &  w/o definitions &11.1&15.0&2.2&4.5&6.0&10.7&8.5&9.4&11.5\\
 \midrule
\multirow{2}{*}{\textbf{\begin{tabular}[c]{@{}l@{}}DeepSeek-v2.5 \end{tabular}}}  &  w/ definitions&13.4&12.2&1.8&-3.8&7.1&8.9&13.2&15.1&14.4\\
 &  w/o definitions &16.8&13.7&0.8&5.0&8.1&13.8&10.9&11.9&10.4\\
 \midrule
\multirow{2}{*}{\textbf{\begin{tabular}[c]{@{}l@{}}Llama 3.1-8B \end{tabular}}}  &  w/ definitions&10.4&14.1&4.9&9.1&9.6&15.3&14.5&12.3&12.4\\
 &  w/o definitions &6.4&6.1&6.9&4.1&5.5&4.0&4.2&5.9&7.1\\
 \midrule
\multirow{2}{*}{\textbf{\begin{tabular}[c]{@{}l@{}}Mixtral 8$\times$7B \end{tabular}}}  &  w/ definitions&7.8&7.4&7.1&-0.5&-4.0&5.6&9.4&6.7&8.3\\
 &  w/o definitions &0.0&-5.1&-2.1&-7.1&3.3&5.7&-3.2&1.0&1.0\\
\bottomrule
\end{tabular}
\end{center}
\caption{The system-level Kendall correlations between human-assigned scores and summary-based evaluation results, comparing scenarios with and without provided criteria definitions.}
\label{table:cri_def} 
\end{table*}

\section{Ablation Studies}

\paragraph{Criteria Definitions.} To investigate the impact of detailed criteria definitions, we conducted experiments both with and without them. As shown in Table \ref{table:cri_def}, providing definitions did not significantly improve performance for powerful models like GPT-4o and DeepSeek-v2.5. However, less powerful models such as Llama 3.1-8B and Mixtral-8$\times$7B show marked improvement with detailed definitions. We attribute this difference to powerful models having already acquired a comprehensive understanding of these aspects through their training, while others have not.
\paragraph{Summary Quality.} 
\begin{table*}[t]
\fontsize{8.5}{11}\selectfont
\begin{center}
\begin{tabular}
{ll|rrrrrrrr|c}
 \toprule
 &  & \multicolumn{1}{c}{\textbf{PLOT}} & \multicolumn{1}{c}{\textbf{CHA}} & \multicolumn{1}{c}{\textbf{WRI}} & \multicolumn{1}{c}{\textbf{WOR}} & \multicolumn{1}{c}{\textbf{THE}} & \multicolumn{1}{c}{\textbf{EMO}} & \multicolumn{1}{c}{\textbf{ENJ}} & \multicolumn{1}{c}{\textbf{EXP}} & \multicolumn{1}{|c}{\textbf{Overall}}  \\
\midrule
\multirow{3}{*}{\textbf{\begin{tabular}[c]{@{}l@{}}GPT-4o\end{tabular}}}  &  Chapter Summaries (4o)&14.8&14.5&8.3&10.7&9.0&15.6&8.2&16.6&14.9\\
 & Overall Summary (4o) &15.3&17.8&4.5&5.0&7.2&12.6&11.8&14.0&13.4\\
 & Overall Summary (4o-mini) &9.7&9.2&-1.6&6.0&8.7&8.5&7.2&14.3&10.1\\
 \midrule
\multirow{3}{*}{\textbf{\begin{tabular}[c]{@{}l@{}}DeepSeek-v2.5 \end{tabular}}} & Chapter Summaries (4o)&14.4&15.0&2.9&1.3&11.5&10.8&10.2&12.2&15.1\\
 & Overall Summary (4o) &13.4&12.2&1.8&-3.8&7.1&8.9&13.2&15.1&14.4\\
 & Overall Summary (4o-mini) &12.3&9.7&-1.5&1.4&6.9&10.6&6.1&12.9&10.8\\
\bottomrule
\end{tabular}
\end{center}
\caption{The system-level Kendall correlations between human-assigned scores and summary-based evaluation results, comparing scenarios that offer different types of summaries.}
\label{tab: summ} 
\end{table*}
To explore how input summaries affect the summary-based evaluations, we conducted experiments with three types: detailed chapter-level summaries, an overall summary, and a lower-quality overall summary (generated by GPT-4o-mini). As shown in Table \ref{tab: summ}, more detailed and higher-quality summaries can be helpful, though not significantly so.
\label{appendix:ablation}
\section{Efficiency Comparison of Different Evaluation Strategies}
The cost comparison of different strategies (\S \ref{sec:lengthy_process}) follows this order: Incremental-Updated > Aggregation-Based > Summary-Based. As mentioned in \S \ref{sec:baselines}, we mitigate inconsistencies in LLM-generated evaluations by averaging 5 results. Thus, the summary-based method is more efficient because it requires only one high-cost summary of the lengthy book, with the 5 evaluations processing only short summaries. In contrast, for aggregation-based evaluation, each of the 5 evaluations must process both the lengthy book and previous summaries. For the incremental-updated method, each evaluation additionally processes the previous evaluations. Taking GPT-4o-based evaluations as an example, the runtime and money cost of 5 evaluations on the test set are shown in Table \ref{tab: efficiency}. In conclusion, while aggregation-based methods perform better than summary-based methods (Table \ref{table:results}), their high cost makes summary-based methods a more cost-effective choice.

\begin{table}[t]
\fontsize{9}{10}\selectfont
\centering
\begin{tabular}{lccc}
\toprule
\textbf{Methods} & \textbf{Input Tokens} & \textbf{Runtime}& \textbf{Cost} \\
\midrule
Summary-based & 3,940K & 770min &\$94\\
Incremental-updated & 12,720K & 4,268min & \$499\\
Aggregation-based & 11,480K & 3,056min & \$416\\
 \bottomrule
\end{tabular}
\caption{The runtime and money cost of different evaluation methods (taking GPT-4o-based evaluations as an example).}
\label{tab: efficiency}
\end{table}

\section{Additional Cases} \label{sec:additional_cases}
\begin{figure*}[t]
    \centering
    \includegraphics[width=0.98\linewidth]{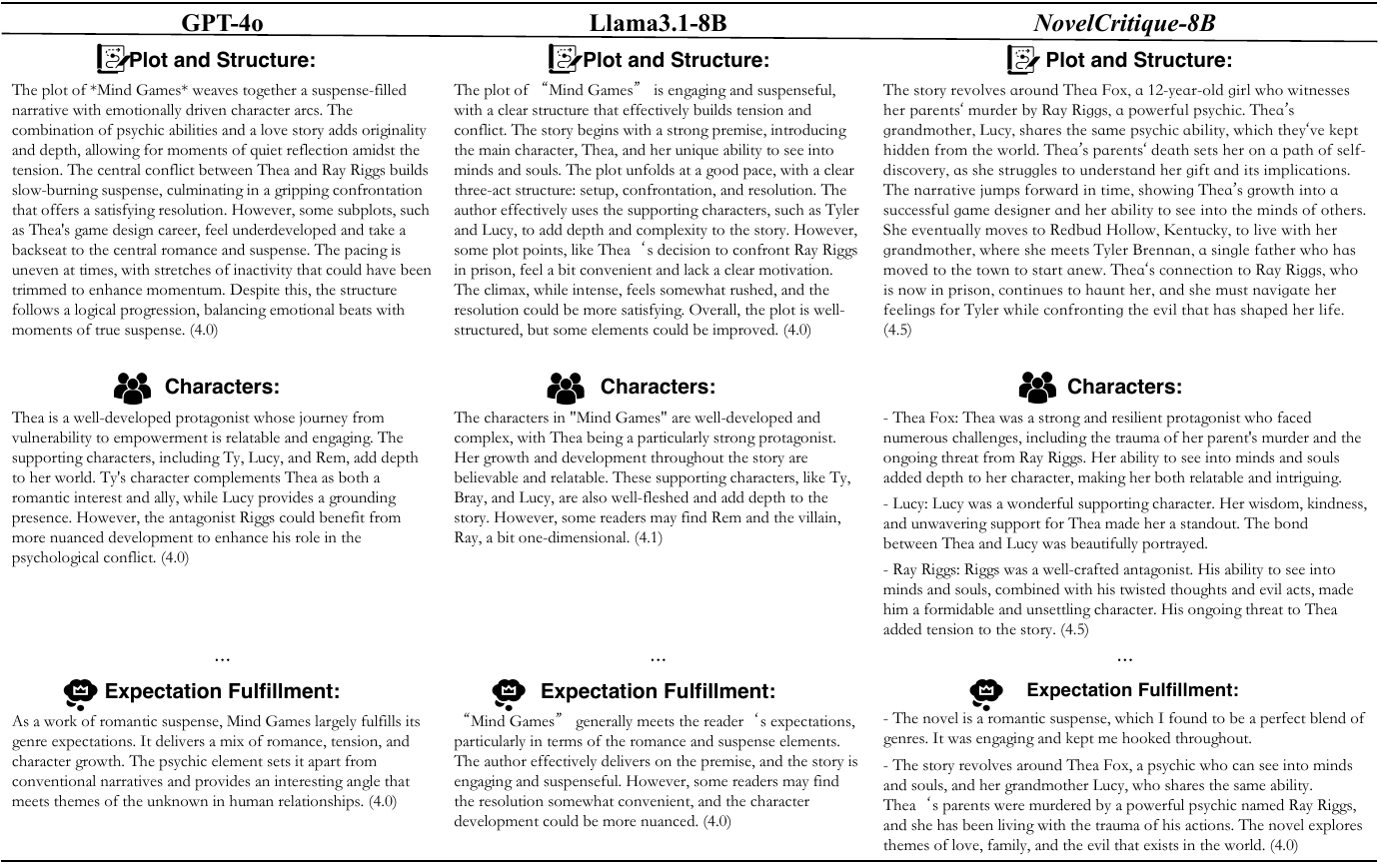}
    \caption{Critiques for ``Mind Games'' (AVG human rating: 4.35).}
    \label{fig:case_1}
\end{figure*}

In Figure \ref{fig:cases}, we present an example of a poorly written story, while Figure \ref{fig:case_1} shows a well-written one. It can be found that NovelCritique offers more concrete and less formulaic comments. 
\section{Prompts}
\subsection{Review Processing}
The review processing prompt is shown in Table \ref{table:review_process}.

\subsection{Instructions of NovelCritique} The instruction of NovelCritique is shown in Table \ref{table:instrction_prompt}.

\subsection{LLM-based Evaluation}\label{appendix:eval_prompts}
The prompt for evaluation is displayed in Table \ref{table:prompt_eval}.

\clearpage

\begin{table*}[t]
\fontsize{7.5}{11}\selectfont
\centering
\begin{tabularx}{\textwidth}{clX}
\toprule
\textbf{Aspects} & \multicolumn{1}{c}{\textbf{Sub-aspects}} & \multicolumn{1}{c}{\textbf{Mapped to normalized criteria}} \\ \midrule
\multirow{6}{*}{\textbf{\begin{tabular}[c]{@{}c@{}}Plot \\ \& Structure\end{tabular}}} & \textbf{Plot Development} & Plot Development [Story Development, Plot Progression, Development], Plot Complexity [Complexity, Plot Twists, Turns, Surprises, Surprise, Plot Predictability, Plot Intensity], Plot Conflict [Conflict, Resolution, Plot Resolution, Conflict Resolution], Pacing [Pace], Plot Clarity \\  [0.6ex]\cline{2-3} 
 & \textbf{Structure} & Plot Structure [Story Structure, Subplots, Timelines, Flow, Narrative Perspective], Plot Elements [Beginning, Middle, Climax, Plot Completion], Plot Coherence [Coherence, Plot Continuity, Consistency, Cohesion], Logic \\ [0.6ex]\cline{2-3} 
 & \textbf{Ending} & Ending [Epilogue, Endings, Closure, Cliffhanger, Conclusion] \\ \midrule
\multirow{11}{*}{\textbf{Characters}} & \textbf{Development} & Character Development [Character Progression], Backstories, Motivations [Character Dynamics], Character Growth, Character Arcs, Character Behavior \\ [0.6ex]\cline{2-3} 
 & \textbf{Characterization} & Characterization [Character Representation], Character Appeal [Character Appreciation, Character Identification], Character Realism, Character Relatability [Character Engagement], Character Depth [Character Complexity], Protagonist [Main Characters, Main Character, Character Focus], Character Introduction \\ [0.6ex]\cline{2-3} 
 & \textbf{Relationships} & Character Relationships [Character Connection, Interactions, Friendships], Character Interactions, Chemistry [Intimacy, Romance and Relationships], Relationship Development [ Family Dynamics, Relationship Dynamics] \\ [0.6ex]\cline{2-3} 
 & \textbf{Diversity} & Character Diversity [Supporting Characters, Secondary Characters, Side Characters, Character Perspectives, Character Perspective, Character Management] \\ \midrule
\multirow{5}{*}{\textbf{\begin{tabular}[c]{@{}c@{}}Writing \\ \& Language\end{tabular}}} & \textbf{Writing Style} & Writing Style [Style, Author's Style, Author's Writing Style, Narrative Style, Author's Skill], Atmosphere, Tone [Author's Tone], Trope [Tropes], Writing Elements [Descriptions, Prose, Imagery, Dialogue, Dialogues] \\ [0.6ex]\cline{2-3} 
 & \textbf{Language} & Grammar [Language Usage, Vocabulary], Fluency [Repetition, Repetitiveness, Author's Reputation] \\ [0.6ex]\cline{2-3} 
 & \textbf{Readability} & Readability [Clarity, Accessibility, Length] \\ \midrule
\multirow{3}{*}{\textbf{Themes}} & \textbf{Exploration/Clarity} & Thematic Elements [Topics, Point of View, Thematic Content, Symbolism, Motifs, Messages], Thematic Exploration [Themes Exploration] \\ [0.6ex]\cline{2-3} 
 & \textbf{Depth} & Thematic Depth [Social Commentary, Educational Value, Influences] \\ \midrule
\multirow{4}{*}{\textbf{\begin{tabular}[c]{@{}c@{}}World-Building \\ \& Setting\end{tabular}}} & \textbf{World-Building} & World-Building [Worldbuilding, World Building], Realism [Cultural Realism, Magic System, Magic, Magic Systems]
\\ [0.6ex]\cline{2-3} 
 & \textbf{Setting} &  Historical Accuracy [Historical Setting, Historical Context, Time Period], Cultural Elements [Cultural References, Cultural Context, Cultural Representation], Technological Elements...\\ \midrule
\multirow{2}{*}{\textbf{Emotional Impact}} & \textbf{Empathy} & Emotional Response [Personal Connection, Emotional Resonance] \\ [0.6ex]\cline{2-3} 
 & \textbf{Depth} & Emotional Depth [Emotional Range] \\ \midrule
\multirow{5}{*}{\textbf{\begin{tabular}[c]{@{}c@{}}Enjoyment\\  \& Engagement\end{tabular}}} & \textbf{Enjoyment} & Enjoyment [Overall Impression, Overall Impact, Excitement, Appeal, Story Enjoyment, Interest, Humor], Anticipation for Future Works [Future Interest, Intrigue,  Anticipation for Future Book, Sequel Potential] \\ [0.6ex]\cline{2-3} 
 & \textbf{Engagement} & Engagement [Reader Engagement, Personal Engagement, Immersion, Story Engagement, Reader Engagement, Overall Experience, Reading Experience] \\ \midrule
\multirow{5}{*}{\textbf{\begin{tabular}[c]{@{}c@{}}Expectation\\  \& Fulfillment\end{tabular}}} & \textbf{Genre} & Genre Expectation, [Genre Appeal, Genre Preference, Audience Appeal, Genre Suitability, Romantic Elements, Thrill, Horror Elements, Fantasy Elements, Historical, Suspense, Tension, Scare Factor, Genre Elements, Genre Classification, Mystery] \\ [0.6ex]\cline{2-3} 
 & \textbf{Premise} & Premise Expectations [Premise Expectation, Story Premise] \\ [0.6ex]\cline{2-3} 
 & \textbf{Title} & Title Relevance \\ \midrule
\multirow{4}{*}{\textbf{Others}} & \textbf{Originality} & Originality [Creativity, Uniqueness, Originality and Creativity, Unique Elements] \\ [0.6ex]\cline{2-3} 
 & \textbf{Content Warnings} & Content Warnings [Warnings, Trigger Warnings] \\ [0.6ex]\cline{2-3} 
 & \textbf{Designment} & Designment [Cover, Cover Design] \\ [0.6ex]\cline{2-3} 
 & \textbf{Personal Bias} & Personal Preference, Personal Experience \\ \bottomrule
 
\end{tabularx}
\caption{The detailed criteria structure.}
\label{table:detailed_criteria}
\end{table*}

\begin{table*}[t]
\fontsize{8.5}{9}\selectfont
\centering
\begin{tabularx}{\textwidth}{X}
\toprule
\textbf{Raw Review Processing} \\
\midrule
As an expert in storytelling, your task is to reformat a reader's review of a full-length novel into a more structured one.
\newline\newline
\textbf{\textit{Objective}}:\newline
- Identify the evaluation aspects discussed in this review.

- Extract the relevant contents that correspond to each identified aspect.

- Organize these contents into a structured format.
\newline\newline
\textbf{\textit{Guidelines}}:\newline
- Maintain the same tone and perspective (e.g., first-person) in the reformatted review, as if you are the original reader.
\newline
- Do not introduce any new information; only reformat the existing review. You may polish the language and improve clarity.
\newline
- Ensure all important viewpoints in the existing review are preserved.
\newline\newline
\textbf{\textit{Output Format:}}\newline
\textbf{Aspects:}\newline
- {List of evaluation aspects identified in the review and the corresponding viewpoints.}

\textbf{Conclusion:}\newline
{Brief summary of the review and a final verdict.}

\textbf{Rating Scores:}\newline
{Any rating scores mentioned in the review.}
\newline\newline
Now, proceed with your task.
\newline\newline
\textbf{Review:}\newline
\{Raw Review\}
\newline\newline
\textbf{Reformatted Review:}

\\
\bottomrule
\end{tabularx}
\caption{The prompt for review processing.}
\label{table:review_process}
\end{table*}

\begin{table*}[t]
\fontsize{9}{10}\selectfont
\centering
\begin{tabularx}{\textwidth}{X}
\toprule
\textbf{Generate Initial Summary} \\
\midrule
Below is the beginning of a lengthy story:
\newline\newline
- - -

\{\}

- - -
\newline\newline
\textbf{\textit{Instructions}}:\newline
Your task is to craft a comprehensive summary of the beginning of a lengthy story, including plot and characters.
\newline\newline
\textbf{\textit{Guidelines}}:\newline
\textbf{1. Plot:}\newline
- Summarize the vital information in the beginning part, such as key events, main conflicts, backgrounds, settings, and characters.\newline
- The story may feature non-linear narratives, flashbacks, or switches between alternate worlds or viewpoints. In such cases, state the time and settings for these narrative shifts.
\newline\newline
\textbf{2. Characters:}\newline
- Introduce the major characters' information, including their profiles and overall experiences.
\newline\newline
\textbf{\textit{Output Format:}}\newline
- - -\newline
\#\#\# Plot Summary:\newline
\{\}
\newline\newline
\#\#\# Characters:\newline
\textbf{Name}:\newline
- \textbf{Profile}: Concisely summarize the character's profile with natural language, including the role (e.g., protagonist), attributes, personality, and relationships within 50 words.\newline
- \textbf{Overall Experience}: Briefly describe the character's overall experience, including motivations, events, and emotional states within 100 words.\newline
- - -\newline
\newline
Now start your summarization:

\\
\bottomrule
\end{tabularx}

\caption{Prompt to generate the initial summary.}
\label{table:prompt_summary_begin}
\end{table*}

\begin{table*}[t]
\fontsize{9}{10}\selectfont
\centering
\begin{tabularx}{\textwidth}{X}
\toprule
\textbf{Update Summary} \\
\midrule
Below is a segment from a story:
\newline\newline
- - -
\newline
\{\}
\newline
- - -
\newline\newline
Below is a summary of the story up until this point:
\newline\newline
- - -
\newline\newline
\#\#\# Plot Summary:
\{\}
\newline\newline
\#\#\# Characters:
\{\}
\newline\newline
- - -
\newline
\newline
\textit{\textbf{Instructions:}}

Your task is to sequentially analyze segments of a lengthy book, updating a comprehensive summary of the entire story, including plot and characters.
\newline\newline
\textit{\textbf{Guidelines:}}\newline
\textbf{1. Plot:}\newline
- First summarize the current segment. You must briefly introduce characters, places, and other major elements if they are mentioned for the first time.\newline
- Update the overall plot summary to include vital information in this segment, related to key events, main conflicts, backgrounds, settings, and characters. \newline
- The story may feature non-linear narratives, flashbacks, or switches between alternate worlds or viewpoints. In such cases, state the time and settings for these narrative shifts.\newline
\newline
\textbf{2. Characters:}\newline
- Update the major characters' information, including their profiles, current experiences, and overall experiences.\newline
- Add new significant characters from this segment to the list.\newline
- If the characters not mentioned in this segment are not crucial to the story's development, remove them from the list of major characters .
\newline\newline
\textit{\textbf{Output Format:}}
\newline- - -\newline
\#\#\# Summary of Current Segment:\newline
\{\}

\#\#\# Overall Plot Summary (within 1000 words):\newline
\{\}

\#\#\# Characters:\newline
\textbf{Name:}\newline
- \textbf{Profile:} Update the character's profile, including the role (e.g., protagonist), attributes, personality, and relationships within 50 words.\newline
- \textbf{Current Experience:} Briefly describe the character's motivations, events, interactions, and emotional states in this segment within 50 words. If the character isn't mentioned, write "Not mentioned".\newline
- \textbf{Overall Experience:} Update the character's overall experience and development throughout the story within 100 words, incorporating the current experience.\newline
- - -
\newline
\newline
Now start your summarization:

\\
\bottomrule
\end{tabularx}
\caption{Prompt to incrementally update the summary of plot and characters.}
\label{table:prompt_summary}
\end{table*}

\begin{table*}[t]
\fontsize{9}{10}\selectfont
\centering
\begin{tabularx}{\textwidth}{X}
\toprule
\textbf{NovelCritique} \\
\midrule
Below are the title, genres, and premise of a book-length story.
\newline

- - -
\newline\newline
\textbf{Title:} \{\}\newline
\newline
\textbf{Genres:} \{\}\newline
\newline
\textbf{Premise:}\newline
\{\}
\newline\newline
- - -
\newline\newline
Below are the plot summary, character analysis, and excerpts from this book-length story.
\newline\newline
- - -
\newline\newline
\textbf{Plot Summary:}\newline
\{\}
\newline\newline
\textbf{Character Analysis:}\newline
\{\}
\newline\newline
\textbf{Excerpts: (Demonstrating Writing and Language):}\newline
\{\}
\newline\newline
- - -
\newline\newline
Write a review of this story based on the criteria below. Then conclude with an overall assessment and assign a final score from 1.0 (lowest) to 5.0 (highest).
\newline\newline
\textbf{\#\#\# Review:}\newline
\textit{/*Specific Aspects*/}
\newline
\textbf{Characters:}\newline
\{Critiques of Characters.\}
\newline
\newline
\textbf{Emotional Impact:}\newline
\{Critiques of Emotional Impact.\}
\newline
\newline
\textbf{\#\#\# Overall Assessment:}\newline
\{Summary of the review.\}
\newline
\newline
\textbf{\#\#\# Score: }X.X

\\
\bottomrule
\end{tabularx}
\caption{Instructions of NovelCritique. Here we display a training sample focusing on two specific aspects: Characters and Emotional Impact.}
\label{table:instrction_prompt}
\end{table*}

\begin{table*}[t]
\fontsize{9}{10}\selectfont
\centering
\begin{tabularx}{\textwidth}{X}
\toprule
\textbf{LLM-based Evaluation} \\
\midrule
Below are the title, genres, and premise of a book-length story.
\newline

- - -
\newline\newline
\textbf{Title:} \{\}\newline
\newline
\textbf{Genres:} \{\}\newline
\newline
\textbf{Premise:}\newline
\{\}
\newline\newline
- - -
\newline\newline
Below are the plot summary, character analysis, and excerpts from this book-length story.
\newline\newline
- - -
\newline\newline
\textbf{Plot Summary:}\newline
\{\}
\newline\newline
\textbf{Character Analysis:}\newline
\{\}
\newline\newline
\textbf{Excerpts:}\newline
\{\}
\newline\newline
- - -
\newline\newline
Please evaluate the story based on the criteria listed below. Provide detailed reviews highlighting both strengths and weaknesses. Assign a score for each aspect on a continuous scale from 0 to 100, where 0 represents the poorest performance and 100 represents the best performance.
\newline\newline
\textbf{\textit{Evaluation Criteria:}}\newline
\#\#\# \textbf{1. Plot and Structure:} \newline
- \textbf{Review:} \newline
- \textbf{Score:} X\newline
- \textbf{Guidelines:} Evaluate the plot development by examining pace, twists, conflicts, and their resolutions. Evaluate the story structure for coherence, logic, and complexity, paying attention to key elements like climax and ending.\newline\newline
\textbf{/*Definitions of other aspects*/}
\newline
\newline
\#\#\# \textbf{Conclusion:} \newline
- \textbf{Overall Assessment:} \newline
- \textbf{Overall Score:} X\newline
- \textbf{Guidelines:} Conclude with an overall assessment of the story, summarizing key insights and assigning an overall score on a continuous scale from 0 to 100.\newline
\newline

Now start your evaluation:

\\
\bottomrule
\end{tabularx}
\caption{Prompt for LLM-based Evaluation. Here we display the prompt for summary-based evaluation, while for others, the difference is to replace the input contents and instructions. For example, in incremental evaluation, we input the current segment, previous summary, and previous evaluations, replacing ``evaluate the story'' as ``update the evaluation''.}
\label{table:prompt_eval}
\end{table*}

\begin{table*}[t]
\fontsize{5.2}{6.8}\selectfont
\centering
\begin{tabular}{llllcc}
\toprule
        \textbf{Title} & \textbf{Author} & \textbf{Genres} & \textbf{Published Date} & \textbf{AVG. Score} & \textbf{Length(Words)} \\ 
        \midrule
            Remedial Magic & Melissa Marr & Fantasy, Romance, LGBT & February 20, 2024 & 2.90 & 88407 \\ 
        Sanctuary of the Shadow & Aurora Ascher & Fantasy, Romance, Romantasy & January 9, 2024 & 2.90 & 96569 \\ 
        We Came to Welcome You & Vincent Tirado & Horror, Thriller, LGBT & September 3, 2024 & 3.12 & 160662 \\ 
        Womb City & Tlotlo Tsamaase & Science Fiction, Horror, Dystopia & January 23, 2024 & 3.20 & 125465 \\ 
        Memory Piece & Lisa Ko & Historical Fiction, Literary Fiction, Contemporary & March 19, 2024 & 3.20 & 80654 \\ 
        The Longest Autumn & Amy Avery & Fantasy, Romance, Adult & January 16, 2024 & 3.21 & 91973 \\ 
        The Girl with No Reflection & Keshe Chow & Fantasy, Young Adult, Romance & August 6, 2024 & 3.28 & 109814 \\ 
        Argylle & Elly Conway & Mystery, Thriller, Mystery Thriller & January 4, 2024 & 3.28 & 113821 \\ 
        Castle of the Cursed & Romina Garber & Fantasy, Horror, Young Adult & July 30, 2024 & 3.30 & 90165 \\ 
        Beach Cute & Beth Reekles & Romance, Young Adult, Contemporary & May 14, 2024 & 3.34 & 88344 \\ 
        All This and More & Peng Shepherd & Science Fiction, Fantasy, Adult & July 9, 2024 & 3.36 & 135548 \\ 
        You Know What You Did & K.T. Nguyen & Thriller, Mystery, Mystery Thriller & April 16, 2024 & 3.36 & 95189 \\ 
        Draw Down the Moon & P.C. Cast & Fantasy, Young Adult, Romance & April 2, 2024 & 3.38 & 80332 \\ 
        Big Time & Ben H. Winters & Science Fiction, Thriller, Mystery & January 9, 2024 & 3.41 & 65522 \\ 
        These Deadly Prophecies & Andrea Tang & Fantasy, Mystery, Young Adult & January 30, 2024 & 3.42 & 70777 \\ 
        Fish Out of Water & Katie Ruggle & Romance, Contemporary, Adult & February 13, 2024 & 3.43 & 75901 \\ 
        Adam and Evie's Matchmaking Tour & Nora Nguyen & Romance, Contemporary, Travel & September 24, 2024 & 3.44 & 129425 \\ 
        The Village Library Demon-Hunting Society & C.M. Waggoner & Fantasy, Mystery, Cozy Mystery & September 24, 2024 & 3.44 & 93691 \\ 
        This Will Be Fun & E.B. Asher & Fantasy, Romance, Adult & October 29, 2024 & 3.45 & 148053 \\ 
        This Ravenous Fate & Hayley Dennings & Fantasy, Lesbian, Vampires & August 6, 2024 & 3.45 & 104419 \\ 
        I Need You to Read This & Jessa Maxwell & Mystery, Thriller, Mystery Thriller & August 13, 2024 & 3.46 & 71906 \\ 
        Lady Macbeth & Ava Reid & Fantasy, Historical Fiction, Retellings & August 13, 2024 & 3.48 & 69586 \\ 
        Every Time I Go on Vacation, Someone Dies & Catherine Mack & Mystery, Cozy Mystery, Mystery Thriller & April 30, 2024 & 3.48 & 76346 \\ 
        Dark Restraint & Katee Robert & Romance, Fantasy, Mythology & August 6, 2024 & 3.49 & 78817 \\ 
        Infinity Alchemist & Kacen Callender & Fantasy, Young Adult, LGBT & February 6, 2024 & 3.49 & 113207 \\ 
        An Academy for Liars & Alexis Henderson & Fantasy, Horror, Gothic & September 17, 2024 & 3.50 & 121043 \\ 
        Love, Lies, and Cherry Pie & Jackie Lau & Romance, Contemporary, Adult & May 7, 2024 & 3.52 & 82144 \\ 
        One Big Happy Family & Jamie    Day & Thriller, Mystery, Mystery Thriller & July 16, 2024 & 3.54 & 106628 \\ 
        Ghost Station & S.A. Barnes & Horror, Science Fiction, Thriller & April 9, 2024 & 3.55 & 105849 \\ 
        Granite Harbor & Peter  Nichols & Mystery, Thriller, Horror & April 30, 2024 & 3.56 & 81080 \\ 
        A Step Past Darkness & Vera Kurian & Thriller, Mystery, Mystery Thriller & February 20, 2024 & 3.56 & 136922 \\ 
        Christa Comes Out of Her Shell & Abbi Waxman & Romance, Contemporary, Chick Lit & April 16, 2024 & 3.57 & 91578 \\ 
        The Getaway List & Emma Lord & Romance, Young Adult, Contemporary & January 23, 2024 & 3.57 & 93259 \\ 
        Last House & Jessica Shattuck & Historical Fiction, Historical, Family & May 14, 2024 & 3.59 & 129626 \\ 
        The Spare Room & Laura Starkey & Romance, Contemporary, Chick Lit & February 6, 2024 & 3.59 & 92741 \\ 
        The Honey Witch & Sydney J. Shields & Fantasy, Romance, LGBT & May 14, 2024 & 3.59 & 98048 \\ 
        The Midnight Feast & Lucy Foley & Mystery, Thriller, Mystery Thriller & June 18, 2024 & 3.59 & 90608 \\ 
        Our Holiday & Louise Candlish & Thriller, Mystery, Crime & July 4, 2024 & 3.60 & 114970 \\ 
        The Ministry of Time & Kaliane Bradley & Science Fiction, Romance, Time Travel & May 7, 2024 & 3.61 & 87170 \\ 
        Sweet Nightmare & Tracy Wolff & Fantasy, Romance, Young Adult & May 7, 2024 & 3.61 & 139647 \\ 
        Hypnotized by Love & Sariah Wilson & Romance, Contemporary, Chick Lit & March 1, 2024 & 3.62 & 79768 \\ 
        The Bad Ones & Melissa Albert & Horror, Fantasy, Young Adult & February 20, 2024 & 3.63 & 87730 \\ 
        The Cautious Traveller's Guide to the Wastelands & Sarah Brooks & Fantasy, Mystery, Historical & June 18, 2024 & 3.63 & 94373 \\ 
        A Friend in the Dark & Samantha M. Bailey & Thriller, Mystery, Mystery Thriller & March 1, 2024 & 3.64 & 67035 \\ 
        The Framed Women of Ardemore House & Brandy Schillace & Mystery, Mystery Thriller, Cozy Mystery & February 13, 2024 & 3.65 & 90841 \\ 
        49 Miles Alone & Natalie D. Richards & Thriller, Young Adult, Mystery & July 2, 2024 & 3.65 & 63689 \\ 
        The Cemetery of Untold Stories & Julia Alvarez & Fantasy, Magical Realism, Historical Fiction & April 2, 2024 & 3.66 & 72967 \\ 
        The Nature of Disappearing & Kimi Cunningham Grant & Thriller, Mystery, Mystery Thriller & June 18, 2024 & 3.66 & 79535 \\ 
        Perfect Little Monsters & Cindy R.X. He & Mystery, Young Adult, Thriller & May 7, 2024 & 3.66 & 68630 \\ 
        Floating Hotel & Grace Curtis & Science Fiction, Mystery, Fantasy & March 19, 2024 & 3.67 & 79893 \\ 
        Ocean's Godori & Elaine U. Cho & Science Fiction, Romance, Queer & April 23, 2024 & 3.67 & 82458 \\ 
        The Honeymoon Affair & Sheila O'Flanagan & Romance & May 9, 2024 & 3.67 & 117284 \\ 
        The House on Biscayne Bay & Chanel Cleeton & Historical Fiction, Mystery, Gothic & April 2, 2024 & 3.68 & 75876 \\ 
        I'm Afraid You've Got Dragons & Peter S. Beagle & Fantasy, Dragons, Humor & May 14, 2024 & 3.69 & 72237 \\ 
        Under Loch and Key & Lana Ferguson & Romance, Fantasy, Paranormal & December 3, 2024 & 3.69 & 105973 \\ 
        Such a Lovely Family & Aggie Blum Thompson & Thriller, Mystery, Mystery Thriller & March 12, 2024 & 3.69 & 94600 \\ 
        Middle of the Night & Riley Sager & Thriller, Mystery, Mystery Thriller & June 18, 2024 & 3.69 & 97895 \\ 
        The Summer She Went Missing & Chelsea Ichaso & Mystery, Thriller, Young Adult & March 1, 2024 & 3.69 & 83624 \\ 
        The Truth According to Ember & Danica Nava & Romance, Contemporary, Indigenous & August 6, 2024 & 3.70 & 91302 \\ 
        The Stardust Grail & Yume Kitasei & Fantasy, Space, Adult & June 11, 2024 & 3.70 & 93381 \\ 
        Beastly Beauty & Jennifer Donnelly & Fantasy, Young Adult, Retellings & May 7, 2024 & 3.70 & 91129 \\ 
        Of Jade and Dragons & Amber Chen & Fantasy, Young Adult, Romance & June 18, 2024 & 3.71 & 103917 \\ 
        Seven Summer Weekends & Jane L. Rosen & Romance, Contemporary, Womens Fiction & June 4, 2024 & 3.72 & 65185 \\ 
        The Vacancy in Room 10 & Seraphina Nova Glass & Thriller, Mystery, Mystery Thriller & April 9, 2024 & 3.73 & 91666 \\ 
        Madwoman & Chelsea Bieker & Mystery, Thriller, Mystery Thriller & September 3, 2024 & 3.73 & 90606 \\ 
        Beautiful Ugly & Alice Feeney & Thriller, Mystery, Mystery Thriller & January 14, 2025 & 3.74 & 84800 \\ 
        Such Charming Liars & Karen M. McManus & Mystery, Young Adult, Mystery Thriller & July 30, 2024 & 3.74 & 91903 \\ 
        The Stars Too Fondly & Emily Hamilton & Science Fiction, Romance, LGBT & June 11, 2024 & 3.76 & 116399 \\ 
        Lies and Weddings & Kevin Kwan & Romance, Contemporary, Chick Lit & May 21, 2024 & 3.76 & 105919 \\ 
        If Something Happens to Me & Alex Finlay & Thriller, Mystery, Mystery Thriller & May 28, 2024 & 3.76 & 61908 \\ 
        Long Island Compromise & Taffy Brodesser-Akner & Historical Fiction, Contemporary, Literary Fiction & July 9, 2024 & 3.77 & 81230 \\ 
        Death at Morning House & Maureen Johnson & Mystery, Young Adult, Mystery Thriller & August 6, 2024 & 3.77 & 85578 \\ 
        The Wren in the Holly Library & K.A. Linde & Fantasy, Romance, Romantasy & June 4, 2024 & 3.78 & 125253 \\ 
        Say you'll be mine: A novel & Naina Kumar & Romance, Contemporary, Adult & January 16, 2024 & 3.79 & 86540 \\ 
        How to Solve Your Own Murder & Kristen Perrin & Mystery, Mystery Thriller, Thriller & March 26, 2024 & 3.79 & 99534 \\ 
        Love Story & Lindsey Kelk & Romance, Contemporary, Chick Lit & July 4, 2024 & 3.81 & 95533 \\ 
        In the Hour of Crows & Dana Elmendorf & Fantasy, Mystery, Magical Realism & June 4, 2024 & 3.81 & 78813 \\ 
        One Last Breath & Ginny Myers Sain & Mystery, Young Adult, Thriller & March 5, 2024 & 3.81 & 93658 \\ 
        A Botanist's Guide to Society and Secrets & Kate Khavari & Mystery, Historical Fiction, Cozy Mystery & June 4, 2024 & 3.81 & 96785 \\ 
        The School Run & Ali Lowe & Thriller, Mystery, Crime & March 5, 2024 & 3.82 & 88185 \\ 
        Five Broken Blades & Mai Corland & Fantasy, Romance, Adult & May 7, 2024 & 3.83 & 114493 \\ 
        The Broken Places & Mia Sheridan & Romance, Mystery, Romantic Suspense & December 1, 2024 & 3.84 & 100338 \\ 
        The Murder After the Night Before & Katy Brent & Mystery, Thriller, Mystery Thriller & February 29, 2024 & 3.85 & 79183 \\ 
        So Let Them Burn & Kamilah Cole & Fantasy, Young Adult, Dragons & January 16, 2024 & 3.85 & 101796 \\ 
        Shelterwood & Lisa Wingate & Historical Fiction, Mystery, Historical & June 4, 2024 & 3.86 & 110366 \\ 
        Dragonfruit & Makiia Lucier & Fantasy, Young Adult, Dragons & April 9, 2024 & 3.87 & 70344 \\ 
        House of Glass & Sarah Pekkanen & Thriller, Mystery, Mystery Thriller & August 6, 2024 & 3.88 & 93373 \\ 
        The Other Valley: A Novel & Scott Alexander Howard & Science Fiction, Fantasy, Time Travel & February 27, 2024 & 3.89 & 92041 \\ 
        Triple Sec & T.J.  Alexander & Romance, LGBT, Queer & June 4, 2024 & 3.89 & 91915 \\ 
        House of Bone and Rain & Gabino Iglesias & Horror, Thriller, Mystery & August 6, 2024 & 3.89 & 104032 \\
        \bottomrule
    \end{tabular}
\caption{Metadata of the test set. (Part 1/2)}
\label{tab:books_info}  
\end{table*}

\begin{table*}[t]
\fontsize{5.2}{6.8}\selectfont
\centering
\begin{tabular}{llllcc}
\toprule
        \textbf{Title} & \textbf{Author} & \textbf{Genres} & \textbf{Published Date} & \textbf{AVG. Score} & \textbf{Length(Words)} \\ 
        \midrule
         
        Voyage of the Damned & Frances  White & Fantasy, Mystery, LGBT & January 18, 2024 & 3.90 & 116476 \\ 
        A Song to Drown Rivers & Ann Liang & Fantasy, Romance, Historical & October 1, 2024 & 3.90 & 93968 \\ 
        The Prisoner’s Throne & Holly Black & Fantasy, Young Adult, Romance & March 5, 2024 & 3.90 & 92811 \\ 
        ASAP & Axie Oh & Romance, Young Adult, Contemporary & February 6, 2024 & 3.91 & 112194 \\ 
        Three Kinds of Lucky & Kim Harrison & Fantasy, Urban Fantasy, Paranormal & March 5, 2024 & 3.91 & 144316 \\ 
        The Queen of Sugar Hill: A Novel of Hattie McDaniel & ReShonda Tate & Historical Fiction, Historical, Race & January 30, 2024 & 3.92 & 160852 \\ 
        Icon and Inferno & Marie Lu & Romance, Young Adult, Mystery & June 11, 2024 & 3.92 & 82204 \\ 
        The Eternal Ones & Namina Forna & Fantasy, Young Adult, Young Adult Fantasy & February 13, 2024 & 3.92 & 115521 \\ 
        The Excitements & C.J. Wray & Historical Fiction, Mystery, Historical & January 30, 2024 & 3.93 & 88857 \\ 
        Hurt Mountain & Angela Crook & Mystery, Thriller, Mystery Thriller & February 27, 2024 & 3.93 & 78349 \\ 
        A Place for Vanishing & Ann Fraistat & Horror, Young Adult, Gothic & January 16, 2024 & 3.96 & 108387 \\ 
        Experienced & Kate Young & Romance, LGBT, Lesbian & June 4, 2024 & 3.98 & 93976 \\
        Still the Sun & Charlie N. Holmberg & Fantasy, Romance, Science Fiction & July 1, 2024 & 4.00 & 85867 \\ 
        The Bright Sword & Lev Grossman & Fantasy, Historical Fiction, Retellings & July 16, 2024 & 4.01 & 184430 \\ 
        Love at First Book & Jenn McKinlay & Romance, Contemporary, Books About Books & May 14, 2024 & 4.01 & 95564 \\ 
        Immortal & Sue Lynn Tan & Fantasy, Romance, Romantasy & January 7, 2025 & 4.03 & 189857 \\ 
        Where Sleeping Girls Lie & Faridah Àbíké-Íyímídé & Mystery, Young Adult, Thriller & March 14, 2024 & 4.03 & 133205 \\ 
        I'm Starting to Worry About This Black Box of Doom & Jason Pargin & Science Fiction, Humor, Mystery & September 24, 2024 & 4.05 & 128001 \\ 
        The Book of Doors & Gareth Brown & Fantasy, Magical Realism, Time Travel & February 13, 2024 & 4.05 & 175207 \\ 
        Deeper Than the Dead & Debra Webb & Mystery, Thriller, Mystery Thriller & August 6, 2024 & 4.05 & 112694 \\ 
        Summer Romance & Annabel Monaghan & Romance, Contemporary, Chick Lit & June 4, 2024 & 4.06 & 77896 \\ 
        Fire and Bones & Kathy Reichs & Mystery, Thriller, Crime & August 6, 2024 & 4.06 & 60889 \\ 
        Here One Moment & Liane Moriarty & Mystery, Contemporary, Thriller & September 10, 2024 & 4.06 & 128430 \\ 
        Reckless & Lauren Roberts & Fantasy, Romance, Romantasy & July 2, 2024 & 4.07 & 97628 \\
        Even If It Breaks Your Heart & Erin Hahn & Romance, Young Adult, Contemporary & February 6, 2024 & 4.08 & 79388 \\ 
        The Hunter & Tana French & Mystery, Ireland, Mystery Thriller & March 5, 2024 & 4.09 & 138237 \\ 
        A Sorceress Comes to Call & T. Kingfisher & Fantasy, Horror, Retellings & August 6, 2024 & 4.10 & 95649 \\ 
        The Missing Witness & Allison Brennan & Mystery, Suspense, Thriller & January 23, 2024 & 4.11 & 98108 \\ 
        Eleven Eleven & Micalea Smeltzer & Romance, Contemporary, Adult & January 24, 2024 & 4.12 & 109791 \\ 
        By Any Other Name & Jodi Picoult & Historical Fiction, Historical, Contemporary & August 20, 2024 & 4.13 & 135244 \\ 
        The Final Act of Juliette Willoughby & Ellery Lloyd & Historical Fiction, Mystery, Thriller & June 11, 2024 & 4.13 & 104445 \\ 
        I Hope This Doesn't Find You & Ann Liang & Romance, Young Adult, Contemporary & February 6, 2024 & 4.16 & 71983 \\ 
        The Day Shelley Woodhouse Woke Up & Laura  Pearson & Romance, Contemporary, Mystery & April 6, 2024 & 4.18 & 78800 \\ 
        The God of the Woods & Liz  Moore & Mystery, Thriller, Mystery Thriller & July 2, 2024 & 4.18 & 118863 \\ 
        The Unquiet Bones & Loreth Anne White & Thriller, Mystery, Mystery Thriller & March 5, 2024 & 4.19 & 93190 \\ 
        The Mercy of Gods & James S.A. Corey & Science Fiction, Space Opera, Fantasy & August 6, 2024 & 4.20 & 114209 \\ 
        The Ghost Orchid & Jonathan Kellerman & Mystery, Crime, Thriller & February 6, 2024 & 4.20 & 64841 \\ 
        Embers in the London Sky & Sarah Sundin & Historical Fiction, Christian Fiction, Romance & February 6, 2024 & 4.21 & 87676 \\ 
        Of Elves and Embers & Elle Madison & Fantasy, Romance, Romantasy & February 6, 2024 & 4.24 & 105599 \\ 
        Disturbing the Dead & Kelley Armstrong & Mystery, Time Travel, Historical Fiction & May 7, 2024 & 4.25 & 114262 \\ 
        Wisteria & Adalyn Grace & Fantasy, Romance, Young Adult & August 20, 2024 & 4.27 & 107341 \\ 
        Think Twice & Harlan Coben & Mystery, Thriller, Mystery Thriller & May 14, 2024 & 4.28 & 79221 \\ 
        Restless Stars & Caroline Peckham & Fantasy, Romance, Romantasy & April 23, 2024 & 4.29 & 308631 \\ 
        The Tainted Cup & Robert Jackson Bennett & Fantasy, Mystery, Mystery Thriller & February 6, 2024 & 4.31 & 110461 \\ 
        The Briar Club & Kate Quinn & Historical Fiction, Mystery, Historical & July 9, 2024 & 4.32 & 188275 \\ 
        The Act of Disappearing & Nathan Gower & Historical Fiction, Mystery, Historical & May 28, 2024 & 4.32 & 100260 \\ 
        Mind Games & Nora Roberts & Romance, Romantic Suspense, Suspense & May 21, 2024 & 4.35 & 127819 \\ 
        Exposed & Brynne Asher & Romance, Romantic Suspense, Suspense & March 26, 2024 & 4.36 & 104163 \\ 
        The Instruments of Darkness & John Connolly & Mystery, Thriller, Horror & May 7, 2024 & 4.41 & 127973 \\ 
        Toxic Prey & John Sandford & Mystery, Thriller, Crime & April 9, 2024 & 4.42 & 86433 \\ 
        All My Secrets & Lynn Austin & Historical Fiction, Christian Fiction, Christian & February 6, 2024 & 4.43 & 120629 \\ 
        The Austrian Bride & Helen Parusel & Historical Fiction, Romance, World War II & January 15, 2024 & 4.43 & 93443 \\ 
        Quicksilver & Callie Hart & Fantasy, Romantasy, Romance & June 4, 2024 & 4.43 & 190189 \\ 
        Heartbeat & Sharon Sala & Romance, Romantic Suspense, Mystery & February 6, 2024 & 4.45 & 101807 \\ 
        Random in Death & J.D. Robb & Mystery, Romance, Crime & January 23, 2024 & 4.47 & 91807 \\ 
        Heart of Frost and Scars & Pam Godwin & Reverse Harem, Dark, Romance & August 26, 2024 & 4.48 & 155855 \\ 
        A Calamity of Souls & David Baldacci & Mystery, Thriller, Crime & April 16, 2024 & 4.49 & 122787 \\ 
        Frank and Red & Matt Coyne & Contemporary, Family, Humor & February 1, 2024 & 4.52 & 100718 \\ 
        James & Percival Everett & Historical, Race, Retellings & March 19, 2024 & 4.53 & 58104 \\ 
        The Women & Kristin Hannah & Historical Fiction, Historical, Adult & February 6, 2024 & 4.62 & 124703 \\  
    \bottomrule

    \end{tabular}
\caption{Metadata of the test set. (Part 2/2)}
\label{tab:books_info_2}  
\end{table*}